%% file: main.tex
\pdfoutput=1
\documentclass[authoryear,3p]{elsarticle}

\input{preamble}

\usepackage{amssymb}
\usepackage{amsmath}
\usepackage{booktabs}
\usepackage{array}
\usepackage{multirow}

\journal{Journal of Hydrology}

\begin{document}

\begin{frontmatter}

\title{C-STRIDE: An Observation-Driven AI Digital Twin for Predicting Basin-Wide Flood Fields from Sparse Stream-Gauge Histories}

\author[label1]{Yanjie Tong}
\author[label1]{Phillip Si}
\author[label1]{Yuan Qiu}
\author[label1]{Peng Chen\corref{cor1}}
\ead{pchen402@gatech.edu}

\affiliation[label1]{organization={School of Computational Science and Engineering, Georgia Institute of Technology},
            addressline={756 West Peachtree Street Northwest},
            city={Atlanta},
            postcode={30308},
            state={GA},
            country={USA}}

\cortext[cor1]{Corresponding author}

\begin{abstract}
Emergency managers need to know where floodwater is, how deep it is, and how it will change over the coming hours across an entire river basin. During a flood, however, real-time measurements come from only a handful of stream gauges, and high-resolution hydrodynamic models are too costly to rerun each time new data arrive or to run as large ensembles. We present C-STRIDE, an observation-driven AI digital twin that turns short records from a few stream gauges, together with terrain and rainfall, into basin-wide maps of water depth and extends these predictions up to a day ahead. It is trained on simulations from a calibrated two-dimensional hydrodynamic model and needs no separate data-assimilation step. In the Des Plaines River basin near Chicago, six gauges inform predictions over 4.2 million 30-m grid cells. Terrain improves the predictions most, rainfall keeps errors from growing over longer horizons, and together they reduce errors by about 40\% compared with gauge records alone. When future rainfall is known, errors remain near 15\% one day ahead, compared with nearly 40\% without rainfall. Given real instead of simulated gauge records, the model shifts its predictions toward the observed hydrographs at three of six gauges without retraining, and it runs about 150 times faster than the hydrodynamic model. These results show how sparse gauges, terrain, and rainfall can be combined into fast, continuously updated flood predictions, a step toward operational flood digital twins that still requires testing with real-time data and rainfall forecasts.
\end{abstract}


\end{frontmatter}

\input{sec_introduction}

\input{sec_dataset}

\input{sec_methodology}

\input{sec_results}

\input{sec_discussion}

\input{sec_conclusion}

\bibliographystyle{elsarticle-harv}
\bibliography{references}

\end{document}

%% file: preamble.tex
\usepackage{xcolor}
\usepackage[hidelinks]{hyperref}
\usepackage{enumitem}

\usepackage{titlesec}
\titleformat{\paragraph}[runin]
  {\normalfont\normalsize\itshape}
  {\theparagraph}{1em}{}

%% file: sec_introduction.tex
\section{Introduction}
\label{sec:introduction}

\subsection{Hydrological digital twins for flood decision support}
\label{sec:motivation}

Flooding threatens lives, infrastructure, and economic activity in cities and regions worldwide~\citep{rentschler2022flood,tellman2021satellite}. Effective response requires information about where water is accumulating, how deep it is, and how inundation may evolve as an event develops. These questions connect local measurements to spatially distributed predictions: a gauge records conditions at one location, while emergency planning requires flood information across roads, neighborhoods, and critical facilities. A useful forecasting system must therefore combine the physical structure of flood dynamics with observations and rapidly updated predictions.

Digital twins provide a framework for this integration. A digital twin is commonly described as a virtual representation of a physical system that is updated with data from that system, has predictive capability, and informs decisions, with two-way interaction between the virtual and physical systems as a central element~\citep{nasem2024digitaltwins}. In hydrology, Digital Twin Earth Hydrology connects Earth-observation products with hydrological modeling to investigate water-cycle conditions and flood-risk scenarios~\citep{brocca2024dthydrology}, observation-coupled hydraulic models support real-time monitoring of drainage systems~\citep{bartos2021pipedream}, and coupled hydrologic--hydrodynamic twins support flood forecasting in data-scarce regions~\citep{rapalo2024floodforecastingdigitaltwin}. For the flood-prediction problem considered here, we organize the required capabilities around three complementary pillars:
\begin{enumerate}[label=(P\arabic*)]
    \item \textbf{Physics-anchored simulation.} A reference model represents the governing hydrodynamics and is evaluated against observations, providing the physical target for a learned surrogate.
    \item \textbf{Fast surrogate replication.} A computationally efficient model reproduces spatial flood evolution, making repeated forecasts and scenario evaluation practical.
    \item \textbf{Observation-driven state estimation.} An observation interface uses incoming measurements to update the estimated basin state and its associated flood field.
\end{enumerate}

Integrating these capabilities is particularly demanding for basin-wide, high-resolution prediction, where sparse measurements must constrain millions of spatial degrees of freedom. In this paper, we develop an observation-driven flood digital twin: a model that provides fast spatial prediction (P2) and observation-driven state estimation (P3), trained on a previously evaluated simulator that supplies the physical reference (P1).
Calibrated uncertainty and decision feedback remain necessary for an operational twin but are outside the present scope; they are discussed in Section~\ref{sec:discussion}.

\subsection{Hydrodynamic models, learned surrogates, and observation integration}
\label{sec:related_dt}

\paragraph{Hydrodynamic models and data assimilation.}
Hydrodynamic solvers remain central to flood modeling because they represent transport, topographic controls, and friction through the shallow water equations. GPU acceleration, including the SynxFlow framework~\citep{xia2017efficient,xia2018new,xia2019full}, makes high-resolution simulations increasingly accessible. Nevertheless, repeated basin-wide simulations remain expensive when forecasts must be updated frequently, ensembles are required, or many scenarios must be evaluated. Connecting such models to observations has a long history in flood forecasting.
 Real-time updating corrects river-model states using gauge data~\citep{madsen2005adaptive}; ensemble Kalman filters update inundation models with spatially distributed water-level measurements~\citep{neal2007flood}; satellite-based assimilation updates river-network and inundation models using observed water levels and flood extents~\citep{garciapintado2015satellite,hostache2018near}; and Pipedream combines hydraulic simulation with Kalman filtering to estimate and forecast drainage-network states~\citep{bartos2021pipedream}. These methods integrate observations through a physics-based model. Here, we examine whether a learned model can map sparse gauge histories directly to the next distributed flood field at lower computational cost.

\paragraph{Learning-based flood surrogates.}
Deep learning has reduced the cost of several flood-modeling tasks~\citep{bentivoglio2022deep}. Convolutional networks emulate flood-depth maps for urban pluvial~\citep{lowe2021uflood,guo2021datadriven} and fluvial~\citep{kabir2020deep} flooding from rainfall, inflow, and topographic inputs. SWE-GNN instead advances hydraulic variables on a computational graph using terrain and current flow conditions, with a propagation rule motivated by finite-volume hydraulics~\citep{bentivoglio2023rapid}. At the scale of individual gauges, long short-term memory (LSTM) networks driven by meteorological forcing rival or exceed conceptual rainfall--runoff models~\citep{kratzert2018rainfall,kratzert2019toward} and underpin operational flood forecasting at large scales~\citep{nevo2022flood,nearing2024global}. These models differ both in what they predict (event-maximum inundation, time-varying hydraulics, or streamflow at gauges) and in what they need at run time, so rainfall-driven, state-driven, and observation-driven surrogates address different problems.

\paragraph{Field reconstruction from sparse sensors.}
A separate line of work reconstructs full spatial fields from a few point sensors. Shallow decoders map instantaneous sensor values to fluid-flow fields~\citep{erichson2020shallow}; Voronoi-tessellation inputs let convolutional networks handle arbitrary sensor layouts~\citep{fukami2021global}; the Senseiver encodes sparse observations with attention and decodes the field at arbitrary query coordinates~\citep{santos2023senseiver}; and SHRED encodes sensor time histories with an LSTM to reconstruct spatiotemporal fields~\citep{williams_sensing_2024}. Coordinate-based decoders conditioned on latent dynamics have also been used to forecast solutions of partial differential equations on irregular geometries~\citep{yin2023continuous,serrano2023operator}. These methods have mostly been demonstrated on fluid-dynamics, climate, or synthetic benchmarks rather than on basin-wide flood fields constrained by real gauge networks.

\paragraph{Conditional latent dynamics for metropolitan floods.}
A particularly relevant reference is the Conditional Latent Dynamics Network (CLDNet)~\citep{si_toward_2026}. It combines rainfall-driven latent dynamics with a coordinate-based decoder conditioned on elevation, slope, and Manning roughness. Pointwise decoding supports irregular watersheds and queries at gauge coordinates without requiring a dense output grid during training. On the Des Plaines River basin, CLDNet demonstrates metropolitan-scale surrogate modeling using a SynxFlow reference evaluated against USGS observations. We reuse this case study and compare against the CLDNet surrogate and against its combination with latent ensemble score filtering (LD-EnSF)~\citep{xiao2026ldensf}, which assimilates the same six gauges.

\paragraph{The observation-to-field problem.}
Emulating a simulator and estimating the current flood state are related but different tasks. A simulator-style surrogate advances a given state under given forcing, whereas an observation-driven model must infer the distributed state from incomplete measurements. A maximum-depth map summarizes event severity but cannot describe when inundation arrives or how quickly it recedes; a time-dependent field can, but requires a representation of the evolving state. Rainfall information remains valuable in either case, particularly for forecasting, but does not replace measurements of the evolving hydraulic response. Conversely, gauge histories constrain recent basin behavior but do not determine future precipitation. An effective architecture should combine these information sources while making their respective contributions explicit. Table~\ref{tab:method_positioning} summarizes representative approaches by the information they require at run time and the output they produce.

\begin{table}[ht]
\centering
\small
\setlength{\tabcolsep}{4pt}
\renewcommand{\arraystretch}{1.18}
\begin{tabular}{@{}>{\raggedright\arraybackslash}p{0.2\linewidth}>{\raggedright\arraybackslash}p{0.27\linewidth}>{\raggedright\arraybackslash}p{0.27\linewidth}>{\raggedright\arraybackslash}p{0.18\linewidth}@{}}
\toprule
\textbf{Approach} & \textbf{Run-time inputs} & \textbf{Output} & \textbf{Spatial representation} \\
\midrule
DTE Hydrology~\citep{brocca2024dthydrology} & Earth observations, hydrological forcing & Water-cycle states and scenarios & Gridded \\
Pipedream~\citep{bartos2021pipedream} & Hydraulic model, sensor data & Drainage-network states and forecasts & 1-D network \\
Inundation data assimilation~\citep{neal2007flood,hostache2018near} & Hydraulic model, gauge or satellite data & Updated water levels and extent & Model grid \\
U-FLOOD~\citep{lowe2021uflood} & Rainfall, terrain & Event-maximum depth & Raster \\
SWE-GNN~\citep{bentivoglio2023rapid} & Current hydraulic state, terrain & Time-evolving hydraulic state & Mesh graph \\
SHRED, Senseiver~\citep{williams_sensing_2024,santos2023senseiver} & Sparse sensor histories or values & Full field & Grid or continuous queries \\
CLDNet~\citep{si_toward_2026} & Rainfall, terrain & Time-evolving depth & Continuous queries \\
\textbf{C-STRIDE (this work)} & Sparse gauge histories, terrain, optional rainfall & Next-step and multi-step depth predictions & Continuous queries \\
\bottomrule
\end{tabular}
\caption{Representative hydrological digital-twin, data-assimilation, flood-surrogate, and sparse-sensing approaches, summarized by the information they require at run time, the output they produce, and how they represent space. Entries describe modeling interfaces rather than a common accuracy benchmark.}
\label{tab:method_positioning}
\end{table}

\subsection{From sparse gauge histories to spatial flood fields}
\label{sec:dt_interface}

Stream gauges provide a practical observation interface because they record the evolving river response at a small number of mainstem and tributary sites. Yet inferring a flood field from these records is underdetermined at any single instant. Similar local depths can occur during different event phases and correspond to different conditions elsewhere in the basin. A temporal history provides information about the direction and rate of change, whereas terrain describes persistent spatial controls that cannot be inferred directly from a few gauge values. Forecasting adds another requirement: the inferred state must evolve consistently with the forcing supplied beyond the observation window.

The STRIDE framework~\citep{stride2026} factorizes sparse-observation reconstruction into a temporal encoder and a coordinate-conditional decoder. The encoder maps an observation history to a compact latent state, and the decoder evaluates the field at a requested spatial coordinate. The rationale comes from delay-embedding theory: if a short history of sensor readings is enough to identify the system state, a network can learn the map from that history to the field. STRIDE makes this precise under a stable delay-observability assumption (Section~\ref{sec:stride_recap}). The assumption is conditional: a finite sensor history need not identify the full state for arbitrary sensor placements or event distributions, so its adequacy here must be examined empirically through field-prediction, gauge, and window-length tests.

Applying this formulation to floods raises three issues. First, local depth depends on elevation, slope, and roughness, so the decoder should receive explicit terrain features alongside the latent basin state. Second, observations and precipitation have different temporal roles: gauge measurements describe the current hydraulic response, whereas rainfall can influence downstream depth after a delay. A model that uses both must keep the direct gauge signal while adding the rainfall information needed to predict what happens next. Third, the evaluation must distinguish agreement with simulator-generated fields from agreement with measured hydrographs. Matching a simulator that has been checked against gauges is useful evidence, but it does not validate every surrogate output against observations.

We therefore develop Conditional STRIDE (C-STRIDE), which combines an LSTM encoder, a terrain-conditioned FMMNN decoder, and a second LSTM that forecasts the gauge readings so that predictions can be extended beyond one step. The encoder receives sparse water-depth histories, augmented with precipitation in the full configuration. The decoder combines the latent state with query coordinates, elevation, slope magnitude, and Manning roughness to predict water depth one step beyond the observation window. For multi-step forecasts, predicted gauge values and the supplied future precipitation are fed back step by step. Terrain-only and forcing-only variants isolate the contributions of the two conditioning sources.

The coordinate decoder suits an irregular watershed, where the active domain occupies only part of the enclosing raster. During training, many locations can be sampled for the same latent basin state, which keeps memory use manageable. At inference, the same representation can be evaluated over an event-specific reduced grid or at a selected set of locations. The same trained model can therefore be tested at both field and gauge scales, without assuming that accuracy at observed sites implies accuracy across the basin.

The trained encoder provides an observation-driven state estimate directly, without an assimilation filter. The estimate is deterministic: the model does not maintain a posterior covariance or produce calibrated uncertainty intervals. Likewise, the coordinate decoder provides flexible spatial evaluation, but querying a finer grid does not by itself establish accuracy beyond the resolution of the reference data.

\subsection{Contributions and evaluation scope}
\label{sec:contributions}

The study evaluates C-STRIDE in the Des Plaines River basin in the greater Chicago area, using $4{,}188{,}840$ active cells at $30$\,m resolution and six USGS gauge locations. The dataset described in Section~\ref{sec:illinois} contains $94$ storm-driven simulations, with $90$ used for training and four held out for testing~\citep{si_toward_2026}. The April 2013 flood, the flood of record at several of these gauges, connects the simulator reference, the learned prediction, and the observed gauge hydrographs for the same event. We organize the study around a question relevant to any flood digital twin: what do sparse gauge histories, terrain, and precipitation each contribute to next-step and multi-step flood prediction? The contributions are:
\begin{enumerate}
    \item[\textbf{(C1)}] \emph{An observation-driven AI digital twin for basin-wide flood prediction.} We adapt STRIDE to combine gauge-history encoding with terrain-conditioned continuous decoding and precipitation-conditioned forecasting. The model provides fast spatial prediction (P2) and observation-driven state estimation (P3), is trained on a simulator evaluated against USGS observations (P1), and can be queried at individual locations or over the basin.
    \item[\textbf{(C2)}] \emph{An empirical separation of spatial and temporal conditioning.} Terrain gives the larger individual improvement in next-step field prediction, while precipitation limits error growth during multi-step forecasting. Joint conditioning reduces four-event mean relative depth error by approximately $41\%$ compared with vanilla STRIDE. Full C-STRIDE reaches $14.48\%$ error at a $24$-h horizon with reference future rainfall, and a perturbation experiment tests sensitivity to that assumption.
    \item[\textbf{(C3)}] \emph{Evaluation across fields, gauges, and observed inputs.} We compare predicted depth and flood extent with SynxFlow on event-specific reduced grids, test hydrograph consistency at mapped gauge cells, replace simulated histories with USGS observations at inference time, and evaluate tributary gauges withheld from the encoder inputs. Additional tests vary history length and train models with reduced, missing, or noisy inputs.
    \item[\textbf{(C4)}] \emph{An assessment of computational cost.} We report field and six-gauge inference times, multi-step evaluation cost, and offline training cost. On the same NVIDIA L40S GPU, next-step prediction of depth fields of a $96$-h event is approximately $150\times$ faster than a SynxFlow simulation, although the two computations differ in scope.
\end{enumerate}

The remainder of the paper is organized as follows. Section~\ref{sec:dataset} describes the study area, data, and problem setup. Section~\ref{sec:methodology} presents the C-STRIDE model, its training, and the evaluation metrics. Section~\ref{sec:results} evaluates next-step prediction, hydrographs, flood extent, forecasting, sensitivity, and computational cost. Section~\ref{sec:discussion} interprets the results, discusses operational implications, and states limitations. Section~\ref{sec:conclusion} concludes.

%% file: sec_dataset.tex
\section{Study area, data, and problem setup}
\label{sec:dataset}

This section describes the simulator that generates the reference data (Section~\ref{sec:synxflow}), defines the prediction problem (Section~\ref{sec:sparse_obs_problem}), and summarizes the Des Plaines River basin dataset and gauge network (Section~\ref{sec:illinois}).

\subsection{High-fidelity simulator for generating the synthetic dataset}
\label{sec:synxflow}

We generated the synthetic training and test datasets using the open-source library SynxFlow~\citep{xia2017efficient, xia2018new, xia2019full, synxflow_software}. The simulator solves the two-dimensional shallow-water equations using a first-order Godunov-type finite-volume method on a uniform rectangular grid. Cells outside the watershed are masked, so the rectangular grid can represent an irregular domain. SynxFlow uses surface reconstruction to preserve well-balanced treatment of bed elevation, the HLLC Riemann solver to compute interface fluxes, and a minmod limiter to reconstruct bed gradients~\citep{xia2018new}. It discretizes the flux and bed-slope terms explicitly and treats the stiff friction term implicitly. The time step is chosen adaptively according to the CFL condition.

The simulator takes as input the initial water depth $h_0$, the initial unit-width discharges $h_0u_0$ and $h_0v_0$ in the $x$- and $y$-directions (where $u_0$ and $v_0$ are depth-averaged velocities), bed elevation $b$, Manning's roughness coefficient $n$, and a spatially varying, discrete-time precipitation series $\{r_t\}$. It outputs the state $(h_t,h_tu_t,h_tv_t)$ at future times $t$.

\subsection{Sparse-observation prediction problem}
\label{sec:sparse_obs_problem}

Let $\Omega\subset\mathbb{R}^2$ denote the physical domain, and let $[0, T]$ be the simulation horizon. We uniformly partition the interval $[0,T]$ into $N_{T}$ subintervals of length $\Delta t= T/N_T$. We consider snapshots of $(h, hu, hv)$ at times $t_k = k\,\Delta t$, for $k = 1,\dots, N_T$. The model is trained to predict only water depth $h(t_k,\xi)$ at locations $\xi\in\Omega$ rather than the unit-width discharges $(hu, hv)$.

We measure water depth through $N_s$ point sensors located at $\{\xi^{(1)},\dots,\xi^{(N_s)}\}\subset\Omega$. The observation vector at time $t_k$ is
\begin{equation}
    y_k := \big(h(t_k,\xi^{(1)}), \dots, h(t_k,\xi^{(N_s)})\big) \in \mathbb{R}^{N_s}.
    \label{eq:obs}
\end{equation}
Other observed quantities, such as water-surface elevation (WSE) or discharge, could be used in the same way with suitable preprocessing and training. We write $y_{k-K:k} = (y_{k-K},\dots,y_k)$ for a window of $K+1$ consecutive observations and $p_k\in\mathbb{R}^{d_p}$ for the precipitation field at time $t_k$. Given a forecast origin $k$, a horizon $m\ge 1$, and a query location $\xi$, the model estimates
\begin{equation}
    \tilde h(t_{k+m},\xi) \;=\; \mathcal{S}_\theta\big(y_{k-K:k},\; p_{k-K:k+m-1},\; \xi,\; \phi(\xi)\big),
    \label{eq:surrogate}
\end{equation}
where $\phi(\xi)$ is a static terrain feature vector (Section~\ref{sec:cstride_arch}). We consider horizons $m=1, 2, \ldots,24$. For $m=1$, we predict water depth at the next time step from historical observations $y_{k-K:k}$ and precipitation $p_{k-K:k}$. For $m\geq 2$, the model additionally requires future precipitation $p_{k+1}, \ldots, p_{k+m-1}$. The default multi-step evaluation supplies the reference future precipitation used to drive SynxFlow; Section~\ref{sec:forecast_results} also tests perturbed rainfall. Operational use would instead require precipitation forecasts.

\subsection{Des Plaines River basin case study}
\label{sec:illinois}

We reuse the Des Plaines River basin (HUC8~07120004) simulation dataset and the six USGS gauges of \citet{si_toward_2026}. SynxFlow solves the two-dimensional shallow-water equations on a fixed $30\,\mathrm{m}$ USGS 3DEP digital elevation model. The computational domain contains $4{,}188{,}840$ active watershed cells within a $5{,}075\times1{,}661$ rectangular grid. NLCD~2021 land cover~\citep{dewitz2023nlcd} determines the Manning coefficients, which are $0.02$ for water-covered cells and $0.05$ for land-covered cells. We refer to~\citet{si_toward_2026} for more details about the simulation setup and data.

The dataset contains $94$ simulations driven by spatially varying NCEP Stage~IV precipitation fields from 2002--2024~\citep{lin2005stageiv}. The rainfall fields come from a broader Midwestern storm archive, so they do not all represent storms that occurred over the Des Plaines basin. At each hourly step, a $39\times13$ precipitation grid supplies $507$ forcing values. Each simulation starts from a shared spun-up flow state and runs for $96\,\mathrm{h}$, producing hourly water depth and discharge fields. We predict water depth only and discard the initial state, retaining the states at original time indices $1,\ldots,96$. Each state is paired with the precipitation at the preceding index ($0,\ldots,95$), so each precipitation input covers the hour before its paired state.

We use $90$ trajectories for training and four for testing, including the April 2013 flood trajectory, which we use for comparison with USGS observations. Our sparse-observation network uses four Des Plaines mainstem gauges and two tributary gauges from \citet{si_toward_2026}. Unless stated otherwise, gauge observations are noise-free simulated water depth sampled at the sensor cells. With $N_s=6$ sensors and the default window of $K+1=12$ stored snapshots, each window contains $72$ gauge values. When precipitation is included, each time step contains $6+507=513$ features, giving an input tensor of size $B\times (K+1) \times513$ for a batch of $B$ windows. We summarize the dataset and sparse-observation configuration in Table~\ref{tab:dataset_summary}. Section~\ref{sec:robustness} also considers missing tributary observations and noisy observations.

\begin{table}[!htb]
\centering
\small
\setlength{\tabcolsep}{6pt}
\renewcommand{\arraystretch}{1.15}
\begin{tabular}{ll}
\toprule
\textbf{Property} & \textbf{Value} \\
\midrule
Domain shape                & Non-rectangular (\texttt{NaN}-masked) \\
Grid size                   & $5{,}075\times1{,}661$ \\
Active in-domain cells      & $4{,}188{,}840$ \\
Spatial resolution          & $30\,\mathrm{m}$ \\
DEM source                  & USGS 3DEP \\
Manning coefficient         & $0.02$ (water) / $0.05$ (land) \\
Precipitation source        & Stage~IV QPE, 2002--2024 \\
Precipitation grid          & $39\times13$ \\
\# precipitation features $d_p$ & $507$ per time step \\
\# trajectories             & $94$ ($90$ train, $4$ test including $1$ USGS comparison) \\
Simulation horizon          & $96\,\mathrm{h}$ \\
Output time interval        & $1\,\mathrm{h}$ \\
Snapshots per trajectory    & $N_T=96$ after discarding the initial state \\
Retained state indices      & $1,2,\ldots,96$ (original time indices) \\
Paired precipitation indices & $0,1,\ldots,95$ (original time indices) \\
Learned output              & Water depth \\
\midrule
Sensor count $N_s$           & $6$ USGS gauges ($4$ mainstem, $2$ tributary) \\
Observed quantity           & Water depth \\
Forecast horizons $m$       & $1$--$24$ steps (reported at time step $1,4,6,12,24$) \\
Simulator evaluation        & $6$ USGS gauges, April 2013 flood \\
\bottomrule
\end{tabular}
\caption{Des Plaines River basin dataset and sparse-observation configuration.}
\label{tab:dataset_summary}
\end{table}

%% file: sec_methodology.tex
\section{The C-STRIDE model}
\label{sec:methodology}

\begin{figure}[ht]
\centering
\includegraphics[width=0.98\columnwidth]{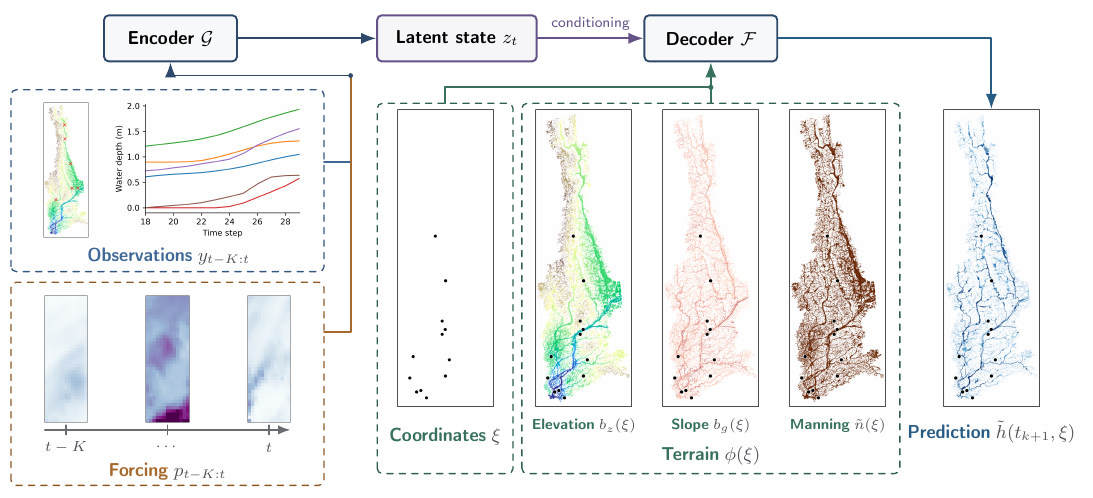}
\vspace{-0.3em}
\caption{Overview of the C-STRIDE architecture.
Sparse observation histories \(y_{k-K:k}\) and precipitation histories \(p_{k-K:k}\) are encoded by \(\mathcal{G}\) into a latent state \(z_k\), which conditions the decoder \(\mathcal{F}\). The decoder is queried at spatial coordinates \(\xi\) together with terrain feature vectors \(\phi(\xi)\), comprising elevation \(b_z(\xi)\), slope \(b_g(\xi)\), and Manning coefficient \(\tilde n(\xi)\), to produce the next-step depth \(\tilde h(t_{k+1},\xi)\). The example panels illustrate sensor observations, precipitation snapshots, query locations, terrain features, and a predicted water-depth field.}
\label{fig:diagram}
\vspace{-0.5em}
\end{figure}

This section develops the Conditional STRIDE (C-STRIDE) model, summarized in Fig.~\ref{fig:diagram}. We start from the original STRIDE factorization (Section~\ref{sec:stride_recap}), introduce terrain conditioning, precipitation input, and the multi-step forecasting scheme (Section~\ref{sec:cstride_arch}), describe training on the basin-wide, high-resolution domain (Section~\ref{sec:training}), and define the inference protocol and evaluation metrics (Section~\ref{sec:metrics}).

\subsection{STRIDE recap: delay-embedded continuous decoding}
\label{sec:stride_recap}

The STRIDE framework~\citep{stride2026} reconstructs a continuous spatiotemporal field from a short window of sparse point measurements by factorizing the reconstruction operator into a temporal encoder $\mathcal G$ and a coordinate-conditional spatial decoder $\mathcal F$:
\begin{equation}
    \tilde h(t_k,\xi) \;=\; \mathcal F\big(\xi, z_k\big), \qquad z_k = \mathcal G(y_{k-K:k}),
    \label{eq:stride_factor}
\end{equation}
where $z_k \in \mathbb{R}^{d_z}$ is a latent state and $K+1$ is the window length. The rationale comes from delay-embedding theory. For autonomous systems, delay-coordinate maps built from a generic observable embed the system's invariant set~\citep{takens1981detecting, mane1981dimension}; \citet{stark1999delay} extended such results to systems driven by deterministic forcing, and \citet{botvinick2025measure} gave a measure-theoretic formulation. Building on these results, \citet{stride2026} show that, when the state on a finite-dimensional parametric invariant set is stably identifiable from $K+1$ delayed observations (stable delay observability), the map in~\eqref{eq:stride_factor} can be approximated to arbitrary accuracy by a recurrent encoder and a coordinate decoder. STRIDE's default instantiation pairs a Long Short-Term Memory (LSTM) encoder~\citep{hochreiter_long_1997} with a modulated Fourier Multi-Component and Multi-Layer Neural Network (FMMNN) decoder~\citep{zhang_fourier_2025} whose frozen random Fourier basis acts as an implicit regularizer.

Two features of the original STRIDE limit its direct use for basin-wide flood prediction. First, the decoder sees only the query coordinate and the latent state, without static terrain features such as elevation, slope, and roughness, and the encoder sees only the sensor values, without external forcing. Flood dynamics are driven by precipitation, which the gauges record only indirectly and with a delay. Delay-embedding results for forced systems treat the forcing as part of the system state~\citep{stark1999delay}, which motivates explicitly supplying the precipitation history as input. Second, the original formulation estimates the field at the end of the observation window and does not specify how to predict beyond it. C-STRIDE trains the encoder--decoder for the next field and uses an auxiliary gauge forecaster for longer horizons.

\subsection{C-STRIDE architecture}
\label{sec:cstride_arch}

\paragraph{Recurrent encoder.}
The encoder $\mathcal G$ is an LSTM that processes the observation window sequentially. In forcing-aware configurations, each observation is concatenated with the paired precipitation field, $\tilde y_i = [y_i, p_i]\in\mathbb{R}^{N_s+d_p}$; otherwise $\tilde y_i = y_i$. Writing $s_i$ for the LSTM hidden state, the update is
\begin{equation}
    s_i \;=\; g\big(s_{i-1}, \tilde{y}_i\big), \qquad i = k-K, \dots, k,
    \label{eq:lstm_input}
\end{equation}
where $g$ is the LSTM cell, and the latent state is $z_k := s_k\in\mathbb{R}^{d_z}$ (the hidden state of the final LSTM layer). Hidden and cell states are reset to zero for each window.

\paragraph{Gauge forecaster and multi-step prediction.}
Unlike~\eqref{eq:stride_factor}, which estimates the field at the end of the window, the C-STRIDE encoder--decoder uses the window ending at $k$ to predict the depth field at $k+1$. A separately trained LSTM, $\mathcal G'$, forecasts the gauge depths one step ahead. Starting from the last observed index $k$, define $y_j^\star=y_j$ for $j\le k$ and $y_j^\star=\hat y_j$ thereafter. At forecast horizon $\ell$,
\begin{equation}
    \hat y_{k+\ell}=\mathcal G'\!\left(\left\{[y_j^\star,p_j]\right\}_{j=k+\ell-K-1}^{k+\ell-1}\right),
    \qquad \ell=1,\dots,M,
    \label{eq:lstm_obs}
\end{equation}
and the encoder recomputes the latent state from the same window,
\begin{equation}
    z_{k+\ell-1}=\mathcal G\!\left(\left\{[y_j^\star,p_j]\right\}_{j=k+\ell-K-1}^{k+\ell-1}\right),
    \qquad \ell=1,\dots,M.
    \label{eq:lstm_pred}
\end{equation}
Precipitation can be included in or omitted from each model independently. Both LSTMs re-encode their fixed-length windows from zero hidden and cell states at every step. The encoder--decoder first predicts the field at $k+\ell$; the predicted gauge depths and the supplied rainfall for that step are then appended to the window, so rainfall at $k+\ell$ first affects the field at $k+\ell+1$.

No new gauge observations are used during a multi-step forecast. This feedback scheme resembles SHRED~\citep{williams_sensing_2024}, except that the latent state is recomputed from a sliding window at each step. The encoder provides an observation-driven state-estimation interface (P3) without maintaining an explicit posterior covariance.

\paragraph{Terrain-conditioned implicit decoder.}
The decoder $\mathcal F$ is a modulated coordinate-based implicit neural representation (INR), conditioned jointly on the latent state and on a static terrain feature vector:
\begin{equation}
    \tilde h(t_{k+\ell}, \xi) \;=\; \mathcal F\big([\gamma(\xi), \phi(\xi)], z_{k+\ell-1}\big), \qquad \ell = 1, \dots, m,
    \label{eq:cond_decoder}
\end{equation}
where $\gamma(\xi)$ is a coordinate feature mapping~\citep{tancik2020fourier}, set to the identity here,
and $\phi(\xi)\in\mathbb{R}^{3}$ collects the terrain features used by CLDNet~\citep{si_toward_2026}:
\begin{equation}
    \phi(\xi) \;=\; \big(b_z(\xi), \; b_g(\xi), \; \tilde n(\xi)\big),
    \label{eq:phi}
\end{equation}
with ground elevation $b_z$, slope magnitude $b_g$, and Manning coefficient $\tilde n$, each scaled to $[-1, 1]$. Because the case study uses two Manning classes, $\tilde n$ acts as a binary water/land indicator. The concatenation $[\gamma(\xi),\phi(\xi)]$ is the decoder input, and a single affine layer maps the latent state to the shift modulations. Each FMMNN layer is defined as in \citet{zhang_fourier_2025}:
\begin{equation}
    \sigma_i(\eta_i) \;=\; A_i \sin\!\big(W_i \eta_i + b_i\big) + c_i + \varphi_i,
\end{equation}
where the frequencies $W_i$ and phases $b_i$ are frozen at random initialization, $A_i$ is a trainable dense mixing matrix, $c_i$ is a trainable bias, and $\varphi_i$ is the latent-derived shift modulation (omitted in the final block).
\citet{stride2026} report that the frozen random Fourier basis acts as an implicit regularizer and trains more stably than fully trainable SIREN networks~\citep{sitzmann_implicit_2020}.

The conditioning in~\eqref{eq:cond_decoder} mirrors that of CLDNet but is attached to an observation-driven rather than a rainfall-driven latent state. The coordinate decoder supplies fast spatial prediction (P2), with queries at selected grid points or other supplied coordinates.

\paragraph{Training objectives.}
For each training window from trajectory $j$, the encoder--decoder receives the observed steps $k-K,\dots,k$ and predicts the field at $k+1$.
At sampled coordinates $\{\xi_q\}_{q=1}^{N_\xi}$, the encoder--decoder parameters $\theta$ minimize normalized-depth mean-squared error,
\begin{equation}
    L_{\mathrm{pred}}(\theta)=\frac{1}{N_\xi}\sum_{q=1}^{N_\xi}\big(\tilde h_\theta(t_{k+1},\xi_q)-h(t_{k+1},\xi_q)\big)^2.
    \label{eq:rec_loss}
\end{equation}
SynxFlow depths provide the physical reference (P1)~\citep{si_toward_2026}. The gauge forecaster $\mathcal G'$ has separate parameters $\theta'$ and minimizes the one-step mean-squared error of the gauge depths,
\begin{equation}
    L_{\mathrm{obs}}(\theta')=\frac{1}{N_sN_{\mathrm{win}}}\sum_{j,k}\big\|\hat y_{k+1}^{(j)}-y_{k+1}^{(j)}\big\|_2^2.
    \label{eq:obs_loss}
\end{equation}
Here, $N_{\mathrm{win}}$ is the number of observation-model training windows. The two models are trained separately.

\subsection{Training strategy and scalability}
\label{sec:training}

Because the decoder~\eqref{eq:cond_decoder} is evaluated pointwise, training cost and memory scale with the number of query points in the loss rather than with the full grid size. As in CLDNet~\citep{si_toward_2026}, we exploit this by training on random subsets of cells.

\paragraph{Spatial subsampling.}
Training and evaluation use, for each event, a reduced grid that keeps only the cells reaching a depth of at least $0.1$\,m in that event's reference simulation. The evaluation domain is therefore defined by each event's reference depths. The four test grids contain $809{,}551$--$993{,}797$ cells. Each training window uses $10^5$ sampled points, with four windows per minibatch. The latent state is shared across all queries within a window.

\paragraph{Optimization.}
The encoder--decoder minimizes~\eqref{eq:rec_loss} using SOAP~\citep{vyas_soap_2025} with initial learning rate $2\times10^{-3}$, momentum coefficients $(0.95,0.95)$, weight decay $0.01$, epsilon $10^{-8}$, and preconditioner updates every ten optimizer steps. The learning rate decreases on training-loss plateaus, with patience of 4 epochs, a factor of 0.4, a relative threshold of 1\%, and a minimum of $10^{-6}$.

The observation models use AdamW with learning rate $10^{-3}$, weight decay $10^{-4}$, batch size $512$, and gradient-norm clipping at $1.0$. The vanilla and terrain-only configurations use a gauge-only forecaster, and the forcing-only and full configurations use a rainfall-conditioned forecaster. Table~\ref{tab:hyperparams} summarizes the architecture and training settings.

\paragraph{Input and output normalization.}
Sensor observations, precipitation, terrain features, and target depths are scaled using training-set minima and maxima to $[-1,1]$. Coordinates retain their stored scaling. The decoder predicts normalized depth, which is mapped back to meters before evaluation; its outputs are not constrained to the normalization interval.

\begin{table}[!htb]
\centering
\small
\setlength{\tabcolsep}{6pt}
\renewcommand{\arraystretch}{1.15}
\begin{tabular}{ll}
\toprule
\textbf{Component} & \textbf{Setting} \\
\midrule
Observation window $K+1$ & $12$ snapshots \\
Encoder $\mathcal G$ & Two-layer LSTM, $256$ hidden units \\
Auxiliary model $\mathcal G'$ & Two-layer LSTM, $128$ hidden units; linear $128\to6$ head \\
Decoder $\mathcal F$ & $8$ FMMNN blocks, width $1024$, rank $128$ \\
Decoder inputs & $2$ coordinates and $3$ terrain features \\
Latent modulation & Affine $256\to896$ shift projection \\
Parameters (full model) & $3.40$\,M total, $2.47$\,M trainable (excluding $\mathcal G'$) \\
Query points per step $N_\xi$ & $10^5$ \\
Minibatch & $4$ windows \\
Optimizer & SOAP, learning rate $2\times10^{-3}$, weight decay $0.01$ \\
Scheduler & ReduceLROnPlateau, patience $4$, factor $0.4$ \\
Epochs & $180$ completed \\
\bottomrule
\end{tabular}
\caption{Architecture and training settings of the full C-STRIDE encoder--decoder and the separately trained gauge forecasters. The four encoder--decoder configurations share the same architecture and optimization settings.}
\label{tab:hyperparams}
\end{table}

\subsection{Digital-twin inference protocol and evaluation metrics}
\label{sec:metrics}

\paragraph{Digital-twin inference protocol.}
At run time, incoming gauge observations update a fixed-length history, which the encoder--decoder uses to predict the depth field one step ahead. Starting from observations through $k$, the model predicts depth at $k+1$ and then generates later fields using predicted gauge values to advance the history window, following~\eqref{eq:lstm_obs}--\eqref{eq:cond_decoder}. At each step, the recurrent states are reset to zero, and the updated window is re-encoded into a latent state that conditions the coordinate decoder. Depth can then be predicted at grid points or gauge coordinates; accuracy, however, is established only on the evaluated reduced grids, and querying finer than $30$\,m adds no information beyond the training resolution.

\paragraph{Evaluation metrics.}
We assess performance along four complementary axes:
\begin{enumerate}
    \item \emph{Field accuracy.} For each evaluation trajectory, we compute the snapshot-wise relative $L_2$ error $\varepsilon_h$ and root-mean-squared error $\mathrm{RMSE}_h$ of water depth over the evaluation cells $\Omega_{\mathrm{eval}}$ (the trajectory-specific reduced grid described in Section~\ref{sec:training}) and average them over the evaluation snapshots. With zero-based indices for the retained sequence, $\mathcal{T}_{\mathrm{eval}}=\{K+1,\dots,N_T-1\}$ corresponds to original state indices $K+2,\dots,N_T$ in Section~\ref{sec:illinois}:
    \begin{equation}
        \varepsilon_h \;=\; \frac{1}{|\mathcal{T}_{\mathrm{eval}}|} \sum_{k\in\mathcal{T}_{\mathrm{eval}}} \frac{\big\|\tilde h(t_k,\cdot)-h(t_k,\cdot)\big\|_{\Omega_{\mathrm{eval}}}}{\big\|h(t_k,\cdot)\big\|_{\Omega_{\mathrm{eval}}}},
        \qquad
        \mathrm{RMSE}_h \;=\; \frac{1}{|\mathcal{T}_{\mathrm{eval}}|} \sum_{k\in\mathcal{T}_{\mathrm{eval}}} \frac{\big\|\tilde h(t_k,\cdot)-h(t_k,\cdot)\big\|_{\Omega_{\mathrm{eval}}}}{\sqrt{|\Omega_{\mathrm{eval}}|}},
        \label{eq:rrmse}
    \end{equation}
    where $\|f\|_{\Omega_{\mathrm{eval}}} = \big(\sum_{\xi\in\Omega_{\mathrm{eval}}} f(\xi)^2\big)^{1/2}$. Because $\varepsilon_h$ is an $L_2$ measure, it is weighted toward deep cells such as channels and reservoirs. Both metrics are computed for each snapshot and then averaged, rather than pooled over space and time.
    \item \emph{Hydrograph metrics at USGS gauges.} For the depth or WSE time series at each mapped gauge cell, we compute the Nash--Sutcliffe efficiency~\citep{nash1970river}, $\mathrm{NSE} = 1 - \sum_k (\tilde h_k - h_k)^2 / \sum_k (h_k - \bar h)^2$; the Kling--Gupta efficiency (2009 formulation)~\citep{gupta2009decomposition}, $\mathrm{KGE} = 1 - \sqrt{(r-1)^2 + (\alpha-1)^2 + (\beta-1)^2}$, where $r$ is the linear correlation, $\alpha$ the ratio of standard deviations, and $\beta$ the ratio of means of the predicted and reference series; and the relative peak-depth error $\varepsilon_{h_{\mathrm{peak}}} = |\max_k \tilde h_k - \max_k h_k| / \max_k h_k$.
    \item \emph{Flood-extent metrics.} For a depth threshold $\tau$, predicted and reference depths are converted to binary masks $\mathbf{1}[h(t,\xi)\ge\tau]$. Summing true positives (TP), false positives (FP), and false negatives (FN) over all evaluation cells and snapshots, we report the critical success index $\mathrm{CSI} = \mathrm{TP}/(\mathrm{TP}+\mathrm{FP}+\mathrm{FN})$~\citep{schaefer1990critical}, precision $\mathrm{TP}/(\mathrm{TP}+\mathrm{FP})$, recall $\mathrm{TP}/(\mathrm{TP}+\mathrm{FN})$, and frequency bias $(\mathrm{TP}+\mathrm{FP})/(\mathrm{TP}+\mathrm{FN})$, which exceeds one when flooding is over-predicted. Following \citet{si_toward_2026}, we use $\tau = 0.5\,\mathrm{m}$.
    \item \emph{Computational cost.} We report the measured evaluation time with its computational scope.
\end{enumerate}
For forecasting, we report $\varepsilon_h$ at the forecast target times as a function of the horizon $m \in \{1, 4, 6, 12, 24\}$.

%% file: sec_results.tex
\section{Results}
\label{sec:results}

This section evaluates the two capabilities that C-STRIDE contributes to the digital twin: fast spatial prediction (P2), assessed through field prediction and flood extent, and observation-driven state estimation (P3), assessed through gauge hydrographs and observed USGS inputs. Physical anchoring (P1) is inherited from the evaluation of SynxFlow against USGS observations by \citet{si_toward_2026} and is not re-evaluated here. We also examine multi-step forecasting, sensitivity to the observation interface, and computational cost.

\paragraph{Compared configurations.}
The full model uses both terrain and precipitation inputs. We compare it with three reduced variants and two CLDNet-based references:
\begin{itemize}
    \item \textbf{Vanilla STRIDE}: no terrain or precipitation input; the STRIDE-FMMNN configuration of \citet{stride2026}, adapted to the observation model of Section~\ref{sec:sparse_obs_problem}.
    \item \textbf{C-STRIDE (terrain)}: terrain input only.
    \item \textbf{C-STRIDE (forcing)}: precipitation input only.
    \item \textbf{C-STRIDE (full)}: terrain and precipitation inputs.
    \item \textbf{CLDNet}: the rainfall-driven CLDNet surrogate~\citep{si_toward_2026}, which uses terrain and rainfall without a gauge-assimilation step.
    \item \textbf{CLDNet + LD-EnSF}: CLDNet coupled with latent ensemble score filtering~\citep{xiao2026ldensf}, which assimilates the gauges without access to the true rainfall.
\end{itemize}
The four STRIDE configurations share the same architecture and training settings and differ only in these inputs.

\begin{table}[!htb]
\centering
\small
\setlength{\tabcolsep}{6pt}
\renewcommand{\arraystretch}{1.15}
\begin{tabular}{lcccc}
\toprule
\textbf{Model} & \textbf{Gauge history} & \textbf{Past rainfall} & \textbf{Future rainfall} & \textbf{Terrain} \\
\midrule
Vanilla STRIDE       & \checkmark & -- & -- & -- \\
C-STRIDE (terrain)   & \checkmark & -- & -- & \checkmark \\
C-STRIDE (forcing)   & \checkmark & \checkmark & \checkmark & -- \\
C-STRIDE (full)      & \checkmark & \checkmark & \checkmark & \checkmark \\
\midrule
CLDNet               & -- & \checkmark & n/a & \checkmark \\
CLDNet + LD-EnSF     & \checkmark & -- & n/a & \checkmark \\
\bottomrule
\end{tabular}
\caption{Information available to each model at run time. Future rainfall is used for forecasting. n/a: not evaluated.}
\label{tab:inputs}
\end{table}

\subsection{Aggregate next-step prediction accuracy}
\label{sec:rmse_results}

Table~\ref{tab:traj_rmse} reports depth prediction errors for the April 2013 flood and the four-event average. Each estimate is one step ahead of a 12-step observation history. The relative error normalizes the depth error by the reference field, and the RMSE gives its magnitude in meters.

\begin{table}[!htb]
\centering
\small
\setlength{\tabcolsep}{5pt}
\renewcommand{\arraystretch}{1.15}
\begin{tabular}{lcccc}
\toprule
\textbf{Model} & \multicolumn{2}{c}{\textbf{April 2013 flood}} & \multicolumn{2}{c}{\textbf{Four-event average}} \\
\cmidrule(lr){2-3}\cmidrule(lr){4-5}
& $\varepsilon_h$ (\%) & $\mathrm{RMSE}_h$ (m) & $\varepsilon_h$ (\%) & $\mathrm{RMSE}_h$ (m) \\
\midrule
Vanilla STRIDE & $13.69$ & $0.0961$ & $18.14$ & $0.0995$ \\
C-STRIDE (terrain) & $9.49$ & $0.0667$ & $14.23$ & $0.0811$ \\
C-STRIDE (forcing) & $13.59$ & $0.0960$ & $15.73$ & $0.0879$ \\
\textbf{C-STRIDE (full)} & $\mathbf{8.38}$ & $\mathbf{0.0589}$ & $\mathbf{10.63}$ & $\mathbf{0.0631}$ \\
\midrule
CLDNet & $13.46$ & $0.0930$ & $16.13$ & $0.0867$ \\
CLDNet + LD-EnSF & $16.99$ & $0.1188$ & $20.88$ & $0.1130$ \\
\bottomrule
\end{tabular}
\caption{Mean snapshot depth errors for the April 2013 flood and the four-event average.}
\label{tab:traj_rmse}
\end{table}

Both conditioning sources improve the aggregate estimates, with terrain alone giving the larger reduction. Full C-STRIDE combines their benefits, reducing the relative error from $18.14\%$ to $10.63\%$ (approximately $41\%$) and yielding the lowest RMSE. Adding rainfall to the terrain-conditioned model further reduces error from $14.23\%$ to $10.63\%$. Full C-STRIDE also improves on the rainfall-driven CLDNet surrogate ($16.13\%$, $0.0867$\,m), although it additionally receives gauge histories. Terrain-only C-STRIDE provides the more relevant comparison with CLDNet + LD-EnSF: its four-event relative error is $14.23\%$ versus $20.88\%$, approximately $32\%$ lower, with RMSE decreasing from $0.1130$ to $0.0811$\,m. Terrain and rainfall therefore provide complementary information: terrain gives the larger individual gain, and rainfall further improves the terrain-conditioned estimate.

The April 2013 results show the same ordering among the STRIDE variants. Terrain conditioning accounts for most of the single-event gain, and the full model reduces relative error from $13.69\%$ to $8.38\%$ and RMSE from $0.0961$ to $0.0589$\,m. Full C-STRIDE also has lower errors than CLDNet ($13.46\%$, $0.0930$\,m), while terrain-only C-STRIDE improves on CLDNet + LD-EnSF ($9.49\%$ versus $16.99\%$).

\begin{figure}[ht]
\centering
\includegraphics[width=0.65\columnwidth]{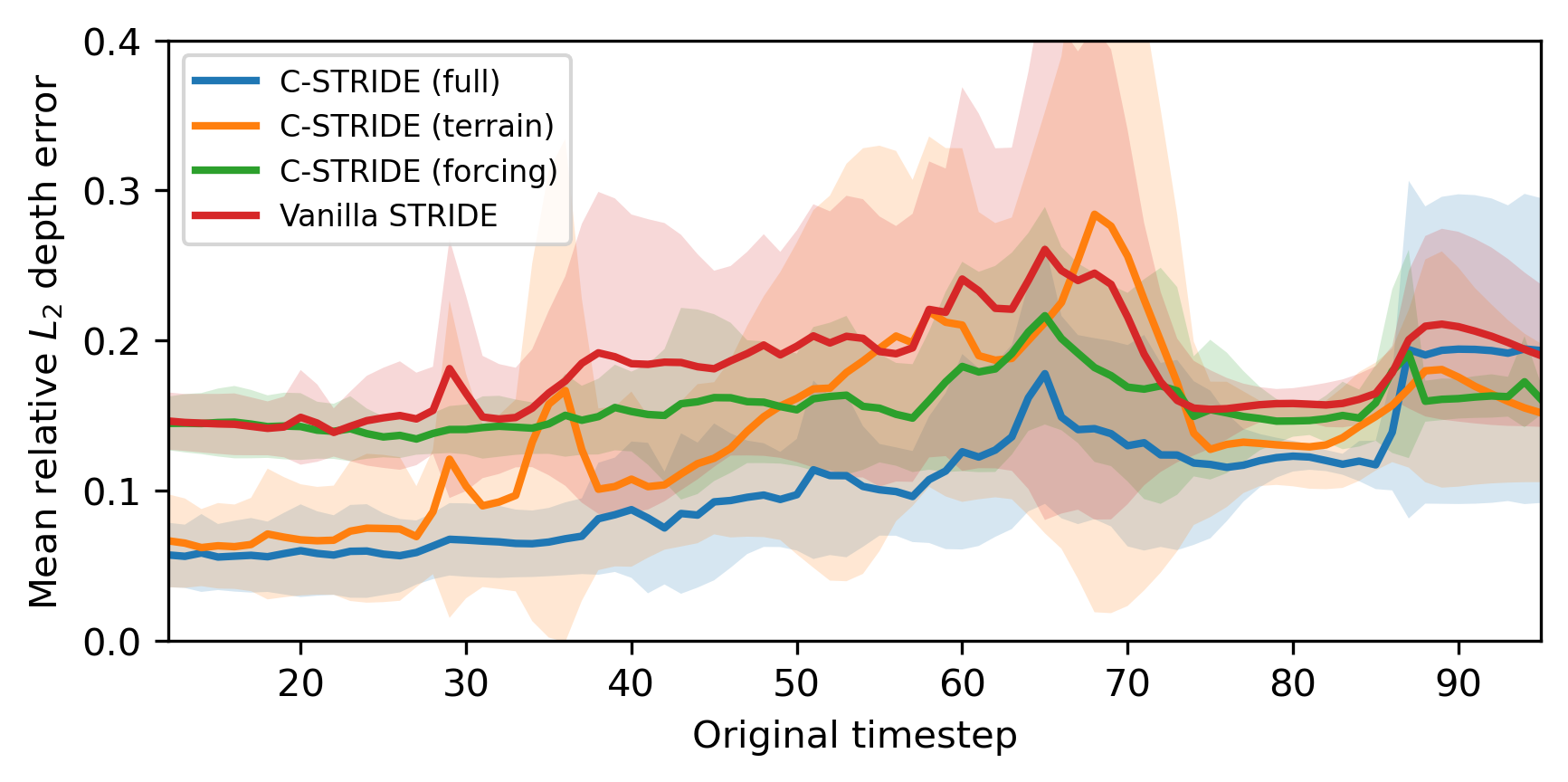}
\caption{Effect of terrain and forcing conditioning on depth prediction. Each curve shows the mean snapshot-wise relative $L_2$ error for one STRIDE configuration. Shading denotes one standard deviation across events.}
\label{fig:ablation_main}
\end{figure}

Figure~\ref{fig:ablation_main} shows the mean prediction error per time step and its variability across events. Full C-STRIDE has the lowest mean error over most of the event, although the curves cross near its end. The separation between configurations varies over time, and wider bands mark periods when prediction difficulty differs more between events.

\begin{figure}[ht]
\centering
\includegraphics[width=0.98\columnwidth]{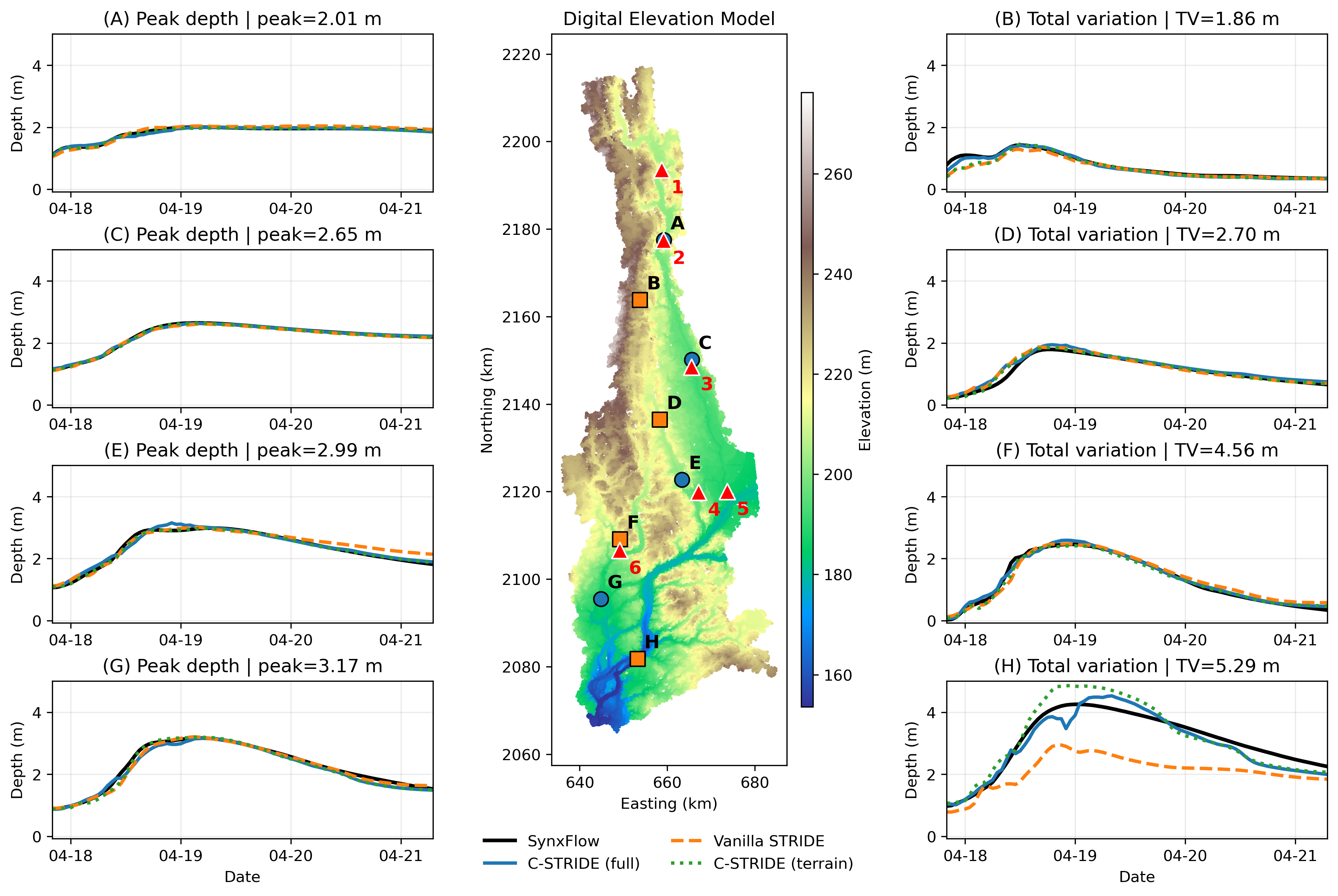}
\vspace{-0.3em}
\caption{April 2013 depth hydrographs at diagnostic locations A--H. Left panels show peak-depth locations (circles), and right panels show high-total-variation locations (squares), selected along alternating northing rows. Curves compare SynxFlow, full C-STRIDE, terrain-only C-STRIDE, and vanilla STRIDE. The center panel shows the DEM, diagnostic locations, and gauges 1--6 (red triangles).}

\label{fig:ablation_vis}
\vspace{-0.5em}
\end{figure}

Aggregate errors can obscure localized behavior. Figure~\ref{fig:ablation_vis} compares full C-STRIDE, terrain-only C-STRIDE, and vanilla STRIDE at eight locations chosen along northing rows, alternating between the largest peak depth, which tests the magnitude of inundation, and the largest temporal total variation, which tests the rise and recession.

Full C-STRIDE and vanilla STRIDE reproduce the broad rise, peak, and recession at the peak-depth locations, with most residual error concentrated around rapid transitions. The clearest difference occurs at location H, the most dynamically variable site. There, vanilla STRIDE underestimates the peak by about $1.31$\,m and misses the recession, whereas full C-STRIDE recovers the overall magnitude and shape but oscillates near the crest and overshoots it by about $0.27$\,m. Terrain-only C-STRIDE overshoots the same crest by $0.61$\,m. Location H lies about 25 km from the nearest gauge, which may leave its rapidly varying local dynamics weakly constrained by the sensor histories. The other high-variation locations show smaller improvements in rising-limb timing and recession.

\begin{figure}[ht]
\centering
\includegraphics[width=0.98\columnwidth]{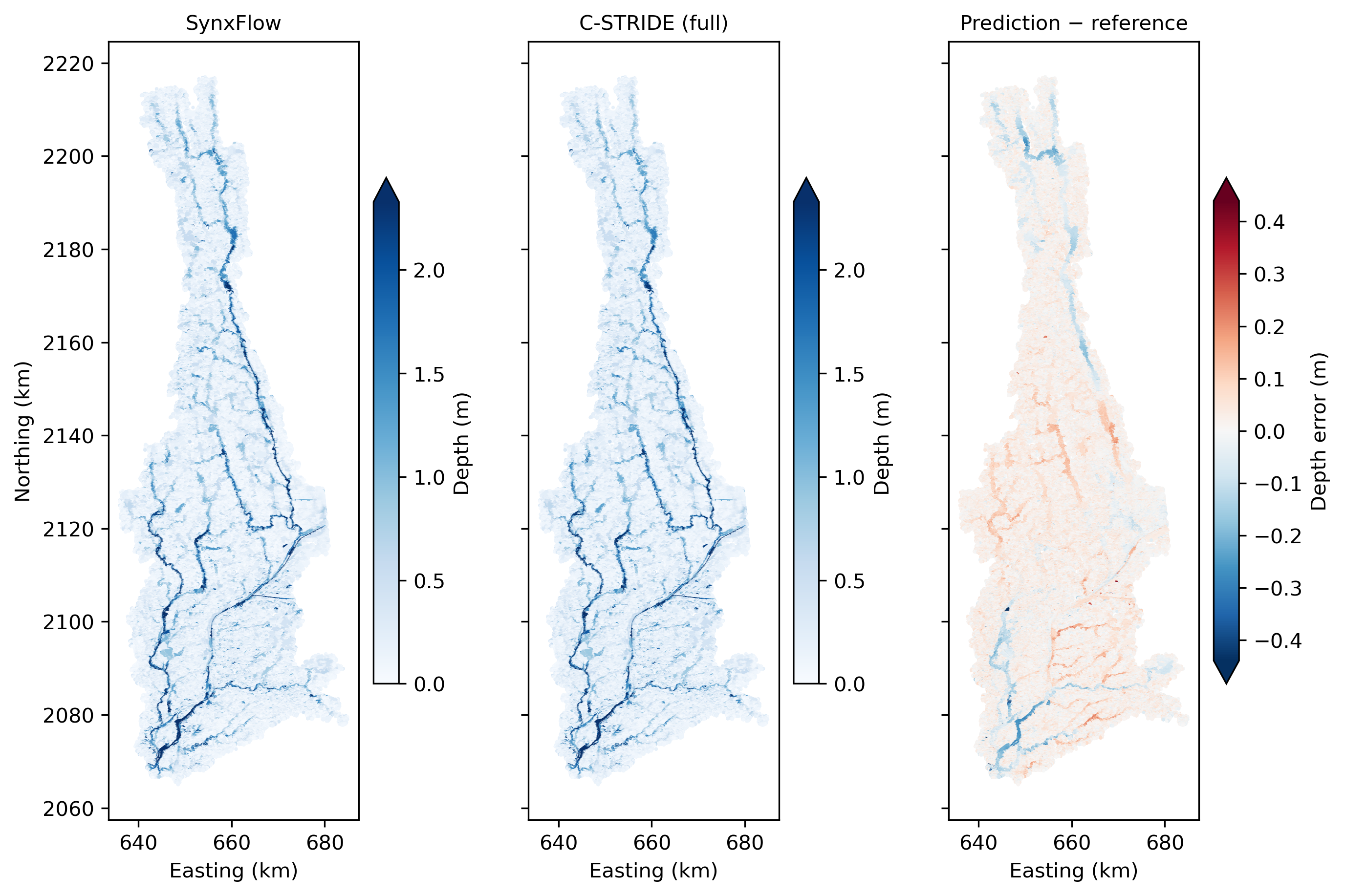}
\caption{Spatial prediction of the April 2013 flood at peak total reference depth. Panels show SynxFlow depth (left), full C-STRIDE depth (center), and prediction minus reference (right) over the evaluation domain.}
\label{fig:error_maps}
\end{figure}

Figure~\ref{fig:error_maps} shows the spatial prediction at peak total reference depth. The snapshot RMSE is $0.0684$\,m, the mean signed error is $0.0045$\,m, and the maximum absolute local error is $1.47$\,m. The full model reproduces the spatial pattern of inundation, and the small positive mean error indicates slight overall over-prediction; larger errors remain at individual locations.

\subsection{Hydrograph fidelity at USGS gauges}
\label{sec:hydrograph_results}

\begin{figure}[ht]
\centering
\includegraphics[width=0.98\columnwidth]{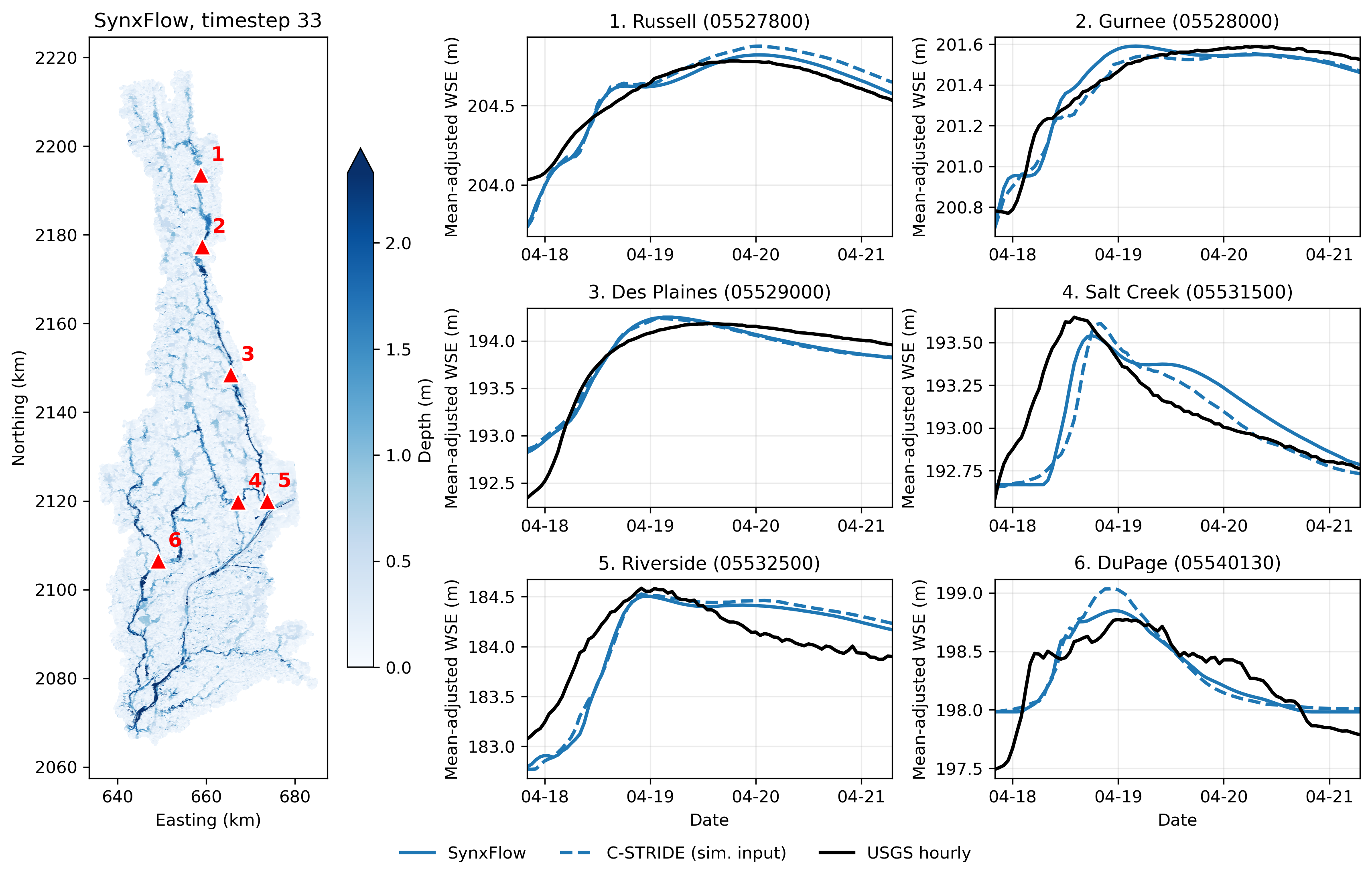}
\vspace{-0.3em}
\caption{Gauge-scale comparison during the April 2013 flood using simulated histories as C-STRIDE inputs. The left panel shows the reference depth and the six gauge locations. Hydrographs compare mean-aligned WSE from full C-STRIDE (blue dashed), SynxFlow (blue solid), and USGS (black solid).}

\label{fig:usgs_val}
\vspace{-0.5em}
\end{figure}

\begin{figure}[ht]
\centering
\includegraphics[width=0.98\columnwidth]{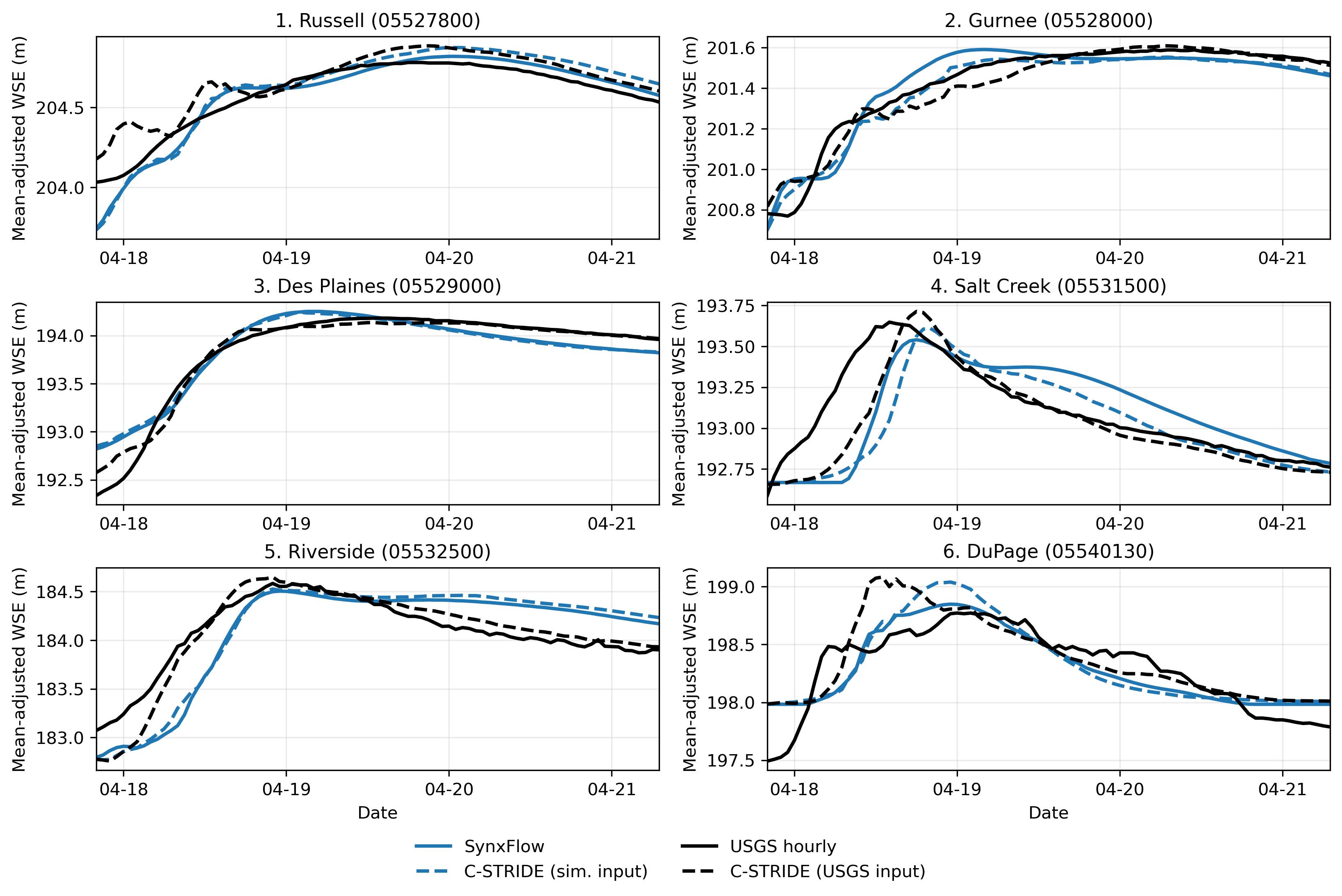}
\vspace{-0.3em}
\caption{Response of full C-STRIDE to replacing simulated gauge histories with USGS observations at inference time, without retraining. Dashed curves show predictions with simulated (blue) or USGS (black) inputs. Solid curves show SynxFlow (blue) and USGS (black). All series use the same retrospective mean alignment, allowing comparison of hydrograph timing and shape.}
\label{fig:usgs_input}
\vspace{-0.5em}
\end{figure}

\begin{figure}[ht]
\centering
\includegraphics[width=0.98\columnwidth]{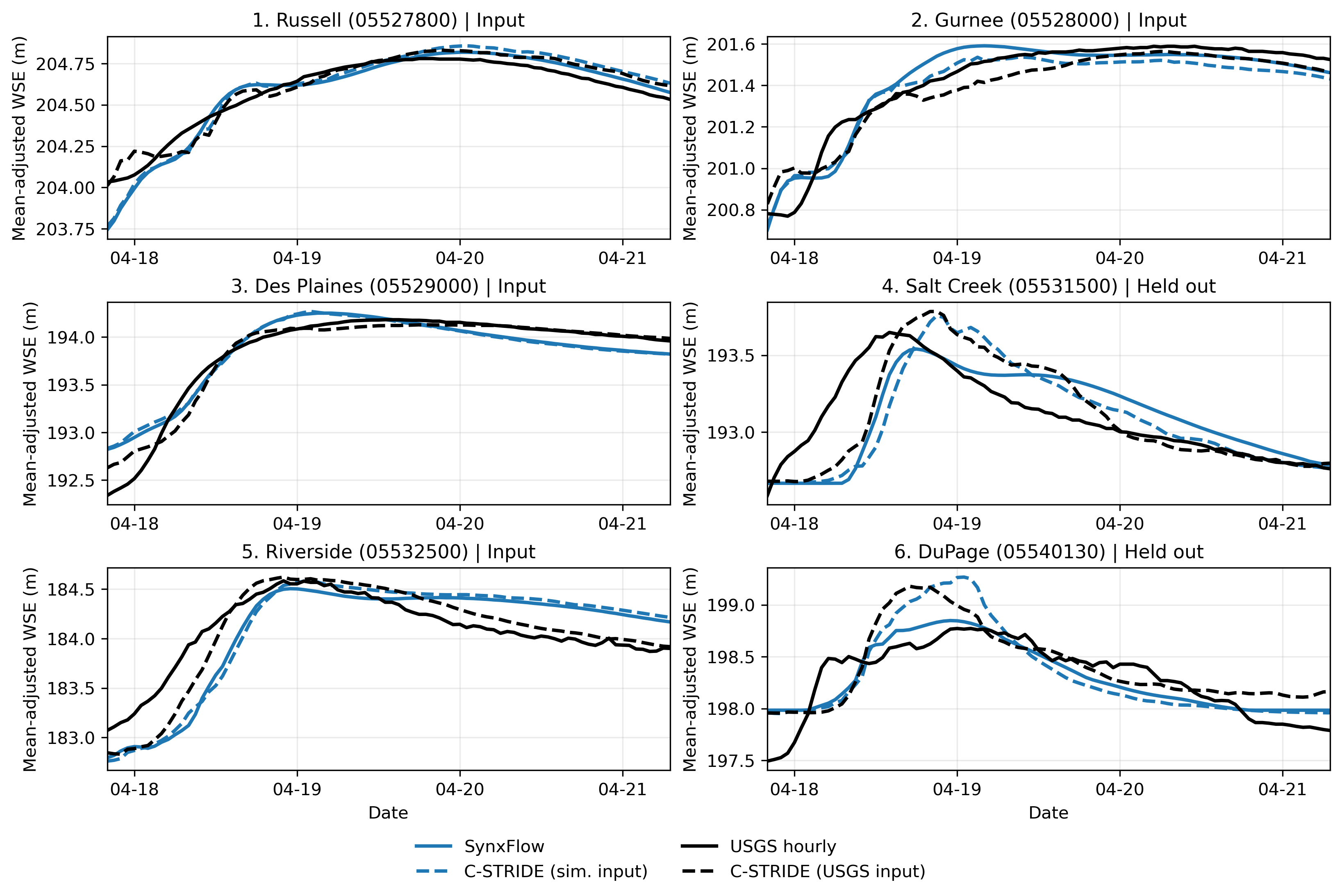}
\vspace{-0.3em}
\caption{Gauge hydrographs from the four-mainstem-input model. Panels marked ``Input'' supply histories to the encoder, while Salt Creek and DuPage, marked ``Held out'', do not supply inputs. Simulated- and USGS-input predictions are compared with SynxFlow and observations using the line styles in Fig.~\ref{fig:usgs_input}.}
\label{fig:usgs_heldout}
\vspace{-0.5em}
\end{figure}

We first compare predicted gauge hydrographs with SynxFlow (Table~\ref{tab:gauge_summary}), which tests whether the predicted field is consistent with the input histories. We then compare predictions with USGS observations, first with simulated and then with observed gauge inputs (Figs.~\ref{fig:usgs_val} and~\ref{fig:usgs_input}), and finally evaluate two tributary gauges withheld from the encoder inputs (Fig.~\ref{fig:usgs_heldout}).

For the USGS comparisons, gauge-specific offsets are computed as the event-wide mean difference between observed and simulated WSE. The same offsets are used to align the WSE series and to convert USGS observations to model-depth inputs. These comparisons therefore assess hydrograph timing and shape after retrospective mean alignment.

\begin{table}[!htb]
\centering
\small
\setlength{\tabcolsep}{5pt}
\renewcommand{\arraystretch}{1.15}
\begin{tabular}{lcccccc}
\toprule
& \multicolumn{2}{c}{\textbf{NSE} $\uparrow$} & \multicolumn{2}{c}{\textbf{KGE} $\uparrow$} & \multicolumn{2}{c}{$\varepsilon_{h_{\mathrm{peak}}}$ (\%) $\downarrow$} \\
\cmidrule(lr){2-3}\cmidrule(lr){4-5}\cmidrule(lr){6-7}
\textbf{Model} & Mean & Median & Mean & Median & Mean & Median \\
\midrule
Vanilla STRIDE           & $0.949$ & $0.937$ & $0.898$ & $0.876$ & $3.10$ & $3.06$ \\
C-STRIDE (terrain)       & $\mathbf{0.976}$ & $\mathbf{0.984}$ & $\mathbf{0.936}$ & $\mathbf{0.952}$ & $\mathbf{2.75}$ & $1.66$ \\
C-STRIDE (forcing)       & $0.931$ & $0.955$ & $0.881$ & $0.880$ & $4.90$ & $\mathbf{0.93}$ \\
\textbf{C-STRIDE (full)} & $0.951$ & $0.960$ & $0.915$ & $0.934$ & $6.33$ & $3.29$ \\
\bottomrule
\end{tabular}
\caption{Gauge hydrograph fidelity relative to SynxFlow during the April 2013 flood. Mean and median are over six gauges, while bold values identify the best configuration.}
\label{tab:gauge_summary}
\end{table}

Terrain-only C-STRIDE gives the best gauge fidelity. It improves NSE over vanilla STRIDE at all six sites and achieves the best mean NSE, KGE, and peak-depth error. Full C-STRIDE improves NSE at five of the six sites, but the gains at Gurnee and Des Plaines are negligible ($\le0.002$), its mean NSE is essentially unchanged ($0.951$ versus $0.949$), and its mean peak-depth error is about twice that of vanilla STRIDE ($6.33\%$ versus $3.10\%$). The larger mean is mainly due to DuPage ($22.04\%$) and Salt Creek ($8.13\%$). By median, which is less sensitive to DuPage, full C-STRIDE improves NSE ($0.960$ versus $0.937$) and KGE over vanilla STRIDE but still has a slightly higher peak error. One possible reason is that the $507$ precipitation values outnumber the $6$ gauge values in the encoder input and may weaken the direct gauge signal, particularly because rainfall affects gauge depth with a delay. Full C-STRIDE is thus best for field prediction, whereas terrain-only conditioning gives the best average fidelity at the gauges.

Figure~\ref{fig:usgs_val} separates surrogate error from simulator-to-observation discrepancy. Full C-STRIDE closely follows SynxFlow at all six sites, while agreement with USGS is strongest for the broad rise and crest at the three upstream mainstem gauges (1--3). The larger timing and recession discrepancies at Riverside (5) and at the tributary gauges (4, 6) are largely shared by C-STRIDE and SynxFlow, indicating that they originate mainly in the simulator rather than in surrogate replication. DuPage (6) is the clearest surrogate-specific exception, where C-STRIDE overshoots the simulated crest by about $0.19$\,m. At both tributary gauges, the SynxFlow WSE is flat at the start of the event while the observed stage is already rising, showing a discrepancy in the early response.

Replacing simulated histories with USGS observations improves NSE from $0.880$ to $0.970$ at Des Plaines, from $0.162$ to $0.603$ at Salt Creek, and from $0.118$ to $0.840$ at Riverside (Fig.~\ref{fig:usgs_input} and Table~\ref{tab:usgs_metrics}). NSE decreases at Russell ($0.780\rightarrow0.731$), Gurnee ($0.944\rightarrow0.932$), and DuPage ($0.588\rightarrow0.567$). At DuPage, the observed-input prediction peaks $11$ hours early and overshoots the observed crest by approximately $0.31$\,m. At Salt Creek and Riverside, the observed histories shift the rise and recession toward the measurements, while at Des Plaines, they improve the late recession. The predictions thus respond to the observed inputs, but not equally well at every gauge.

\begin{table}[!htb]
\centering
\small
\setlength{\tabcolsep}{4pt}
\renewcommand{\arraystretch}{1.15}
\begin{tabular}{lcccccc}
\toprule
& \multicolumn{2}{c}{\textbf{SynxFlow}} & \multicolumn{2}{c}{\textbf{C-STRIDE (sim.\ input)}} & \multicolumn{2}{c}{\textbf{C-STRIDE (USGS input)}} \\
\cmidrule(lr){2-3}\cmidrule(lr){4-5}\cmidrule(lr){6-7}
\textbf{Gauge} & NSE & $\Delta t_{\mathrm{peak}}$ (h) & NSE & $\Delta t_{\mathrm{peak}}$ (h) & NSE & $\Delta t_{\mathrm{peak}}$ (h) \\
\midrule
Russell (05527800) & $0.883$ & $+6$ & $0.780$ & $+7$ & $0.731$ & $+3$ \\
Gurnee (05528000) & $0.893$ & $-25$ & $0.944$ & $+2$ & $0.932$ & $+2$ \\
Des Plaines (05529000) & $0.889$ & $-10$ & $0.880$ & $-11$ & $0.970$ & $0$ \\
Salt Creek (05531500) & $0.163$ & $+4$ & $0.162$ & $+6$ & $0.603$ & $+4$ \\
Riverside (05532500) & $0.182$ & $+1$ & $0.118$ & $0$ & $0.840$ & $0$ \\
DuPage (05540130) & $0.712$ & $-1$ & $0.588$ & $-1$ & $0.567$ & $-11$ \\
\midrule
Held out: Salt Creek & $0.163$ & $+4$ & $0.004$ & $+7$ & $0.346$ & $+7$ \\
Held out: DuPage & $0.712$ & $-1$ & $0.388$ & $+1$ & $0.387$ & $-7$ \\
\bottomrule
\end{tabular}
\caption{Agreement with USGS mean-aligned WSE. Positive peak-time errors indicate late predictions. The first maximum defines peak time. At Gurnee, the observed crest is nearly flat, so peak-time errors there are sensitive to small fluctuations. The last two rows use gauges withheld from encoder inputs.}

\label{tab:usgs_metrics}
\end{table}

\paragraph{Held-out tributary gauges.}
The four-mainstem-gauge model does not use Salt Creek or DuPage as inputs, although both locations remain in the simulated training fields. The response at these two held-out tributaries is mixed (Fig.~\ref{fig:usgs_heldout}). With USGS inputs, Salt Creek NSE improves from $0.004$ to $0.346$, but its peak remains $7$ hours late. At DuPage, NSE is essentially unchanged ($0.388$ versus $0.387$), and the peak shifts from $1$ hour late to $7$ hours early. Observed mainstem histories can therefore improve a tributary prediction without that tributary's own record, but not reliably.

\subsection{Flood-extent metrics}
\label{sec:extent_results}

Beyond pointwise depth accuracy, a flood digital twin must identify the spatial extent of decision-relevant inundation. Following \citet{si_toward_2026}, Table~\ref{tab:extent_metrics} reports CSI, precision, recall, and frequency bias at the $\tau=0.5\,\mathrm{m}$ depth threshold for the 2013 event, aggregated over all valid snapshots and evaluation cells.

\begin{table}[!htb]
\centering
\small
\setlength{\tabcolsep}{6pt}
\renewcommand{\arraystretch}{1.15}
\begin{tabular}{lcccc}
\toprule
\textbf{Model} & \textbf{CSI} $\uparrow$ & \textbf{Precision} $\uparrow$ & \textbf{Recall} $\uparrow$ & \textbf{Bias} ($\to1$) \\
\midrule
Vanilla STRIDE           & $0.888$ & $0.947$ & $0.935$ & $0.987$ \\
C-STRIDE (terrain)       & $0.925$ & $0.959$ & $0.963$ & $1.004$ \\
C-STRIDE (forcing)       & $0.894$ & $0.951$ & $0.938$ & $0.986$ \\
\textbf{C-STRIDE (full)} & $\mathbf{0.933}$ & $\mathbf{0.963}$ & $\mathbf{0.967}$ & $1.004$ \\
\midrule
CLDNet                   & $0.871$ & $0.926$ & $0.936$ & $1.011$ \\
CLDNet + LD-EnSF         & $0.843$ & $0.901$ & $0.929$ & $1.032$ \\
\bottomrule
\end{tabular}
\caption{Flood-extent metrics for the April 2013 flood at a $0.5$\,m depth threshold. Frequency bias above one indicates over-prediction.}
\label{tab:extent_metrics}
\end{table}

Full C-STRIDE has the highest CSI, precision, and recall, with CSI reaching $0.933$. For this single event, terrain conditioning accounts for most of the improvement in flood extent, while precipitation alone has only a modest effect. Adding precipitation to the terrain-conditioned model yields the best overall inundation mask.
Precision and recall show the same pattern. Vanilla STRIDE slightly under-predicts flooding (bias $0.987$), and terrain conditioning mainly corrects these omissions without adding many false alarms. Full C-STRIDE is nearly unbiased, whereas both CLDNet references slightly over-predict. Because cells that stay wet throughout the event, such as river channels, count as hits in these pooled scores, the high absolute CSI values partly reflect easy cells~\citep{stephens2014problems}.

\subsection{Multi-step forecast skill}
\label{sec:forecast_results}

Each multi-step forecast predicts $24$ successive hourly depth fields without new gauge observations. We evaluate $40$ forecasts, from $10$ randomly chosen origins per event, using the same origins for all configurations. The rainfall-conditioned configurations receive the observed future rainfall. Because predicted gauge values replace the unavailable observations as the window advances, errors can accumulate in both the gauge forecaster and the field prediction, so the results test the complete forecasting procedure.

\begin{table}[!htb]
\centering
\small
\setlength{\tabcolsep}{5pt}
\renewcommand{\arraystretch}{1.15}
\begin{tabular}{lccccc}
\toprule
\textbf{Model} & $m=1$ & $m=4$ & $m=6$ & $m=12$ & $m=24$ \\
\midrule
Vanilla STRIDE & $17.26\pm8.72$ & $18.15\pm9.12$ & $21.05\pm11.35$ & $27.69\pm13.42$ & $38.14\pm12.63$ \\
C-STRIDE (terrain) & $11.72\pm9.75$ & $13.49\pm9.84$ & $17.42\pm12.30$ & $26.88\pm14.81$ & $38.74\pm13.68$ \\
C-STRIDE (forcing) & $14.70\pm4.29$ & $15.33\pm4.53$ & $15.76\pm4.52$ & $16.86\pm5.07$ & $18.28\pm4.26$ \\
\textbf{C-STRIDE (full)} & \boldmath$8.06\pm5.09$ & \boldmath$8.72\pm4.97$ & \boldmath$9.38\pm4.75$ & \boldmath$11.75\pm5.88$ & \boldmath$14.48\pm4.67$ \\
Full, rainfall mismatch $\alpha=1$ & $8.06\pm5.09$ & $8.74\pm5.01$ & $9.44\pm4.85$ & $12.05\pm6.16$ & $14.72\pm4.63$ \\
Full-field persistence & $2.73\pm3.85$ & $9.56\pm10.43$ & $13.02\pm12.92$ & $21.74\pm16.21$ & $36.37\pm17.79$ \\
\bottomrule
\end{tabular}
\caption{Mean $\pm$ standard deviation of relative depth error (\%) at forecast horizons $m \in \{1, 4, 6, 12, 24\}$ across 40 forecasts. Persistence uses the complete preceding reference field.}
\label{tab:forecast_horizon}
\end{table}

Full C-STRIDE yields the lowest mean error among the four unperturbed configurations for each horizon (Table~\ref{tab:forecast_horizon}). Terrain improves early forecasts, but its advantage over vanilla STRIDE has disappeared by $24$ hours. Rainfall conditioning limits this error growth: forcing-only C-STRIDE reaches $18.28\%$, while the full model reaches $14.48\%$, about $62\%$ below vanilla STRIDE. Full-field persistence has lower error at horizon one ($2.73\%$ versus $8.06\%$), while full C-STRIDE is better at the reported horizons from 4 onward and reaches $14.48\%$ versus $36.37\%$ at horizon 24. Persistence uses the complete previous reference field, unlike the sparse-input learned models.

The one-step errors in Tables~\ref{tab:forecast_horizon} ($8.06\%$) and~\ref{tab:traj_rmse} ($10.63\%$) use different evaluation snapshots. The former averages over $40$ randomly selected forecast origins (ten per event), restricted to time steps $12$--$72$ to accommodate the $24$-step forecast, whereas the latter averages over all $336$ valid snapshots at time steps $12$--$95$ across the four events. For the gauge forecasts alone, the error of the gauge-only forecaster rises from $1.27\%$ at the first step to $42.63\%$ at $24$\,h, compared with $9.30\%$ and $16.48\%$ for the rainfall-conditioned forecaster. Rainfall therefore helps long-horizon gauge forecasting but not the first step.

\paragraph{Sensitivity to future precipitation.}
To test sensitivity to the reference-rainfall assumption, we perturb the normalized future precipitation $p_{k+\ell}\in[-1,1]^{d_p}$ at future step $\ell\in\{1,\dots,M\}$ as
\begin{equation}
\begin{aligned}
    p^{\mathrm{mis}}_{k+\ell}
    &= \operatorname{clip}_{[-1,1]}\!\left(
        p_{k+\ell}
        + \alpha\frac{\ell}{M}\,\sigma_{k+\ell}\epsilon_{k+\ell}
      \right), \\
    \sigma_{k+\ell}
    &= \operatorname{std}\!\left[p_{k+\ell}\right],
    \qquad
    \epsilon_{k+\ell}\sim\mathcal N(0,I).
\end{aligned}
    \label{eq:precip_mismatch}
\end{equation}
The perturbation grows linearly with lead time and scales with the spatial variability of the true rainfall field. We use $\alpha=1$ (``Noisy'') and apply the same realization to all components that consume future precipitation. Observed histories are unchanged. Because rainfall is appended after each field prediction, the perturbation first affects the following predicted field.

\begin{figure}[ht]
\centering
\includegraphics[width=0.65\columnwidth]{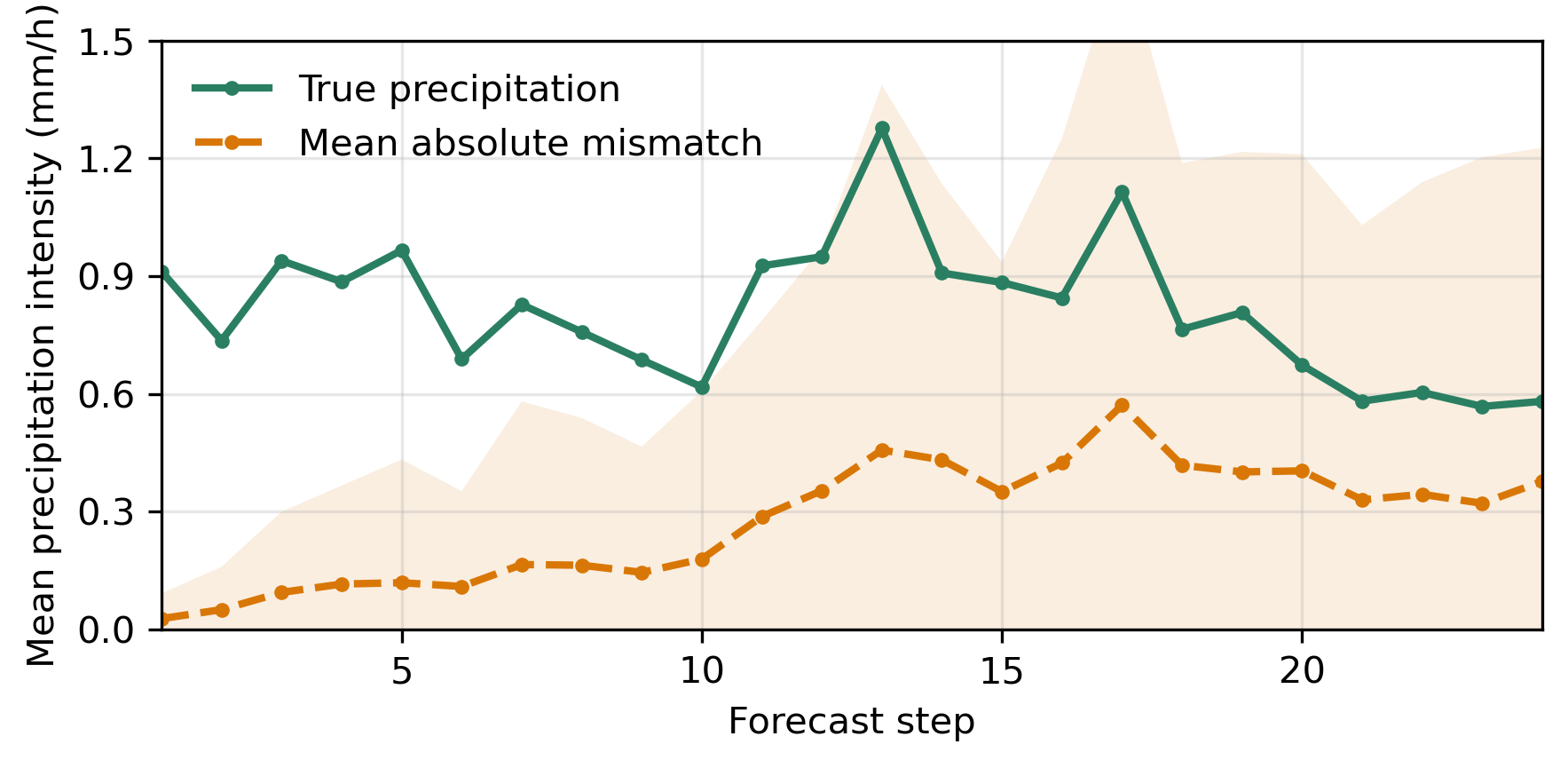}
\vspace{-0.3em}
\caption{Mean rainfall intensity and mean absolute mismatch for $\alpha=1$ over the forecast horizon. The solid curve shows true precipitation and the dashed curve shows the imposed mismatch, averaged equally over the $507$ rainfall values and then over forecasts. Shading denotes one standard deviation of the mismatch across forecasts.}

\label{fig:precip_mismatch_intensity}
\vspace{-0.5em}
\end{figure}

Figure~\ref{fig:precip_mismatch_intensity} compares the mean rainfall intensity with the mean absolute mismatch. Averaged over the $507$ rainfall values (not area-weighted), the true intensity is $0.812$\,mm/h, the absolute mismatch is $0.277$\,mm/h, and the signed mismatch is $+0.072$\,mm/h. The band shows how the size of the perturbation varies between forecast origins.

\begin{figure}[ht]
\centering
\includegraphics[width=0.65\columnwidth]{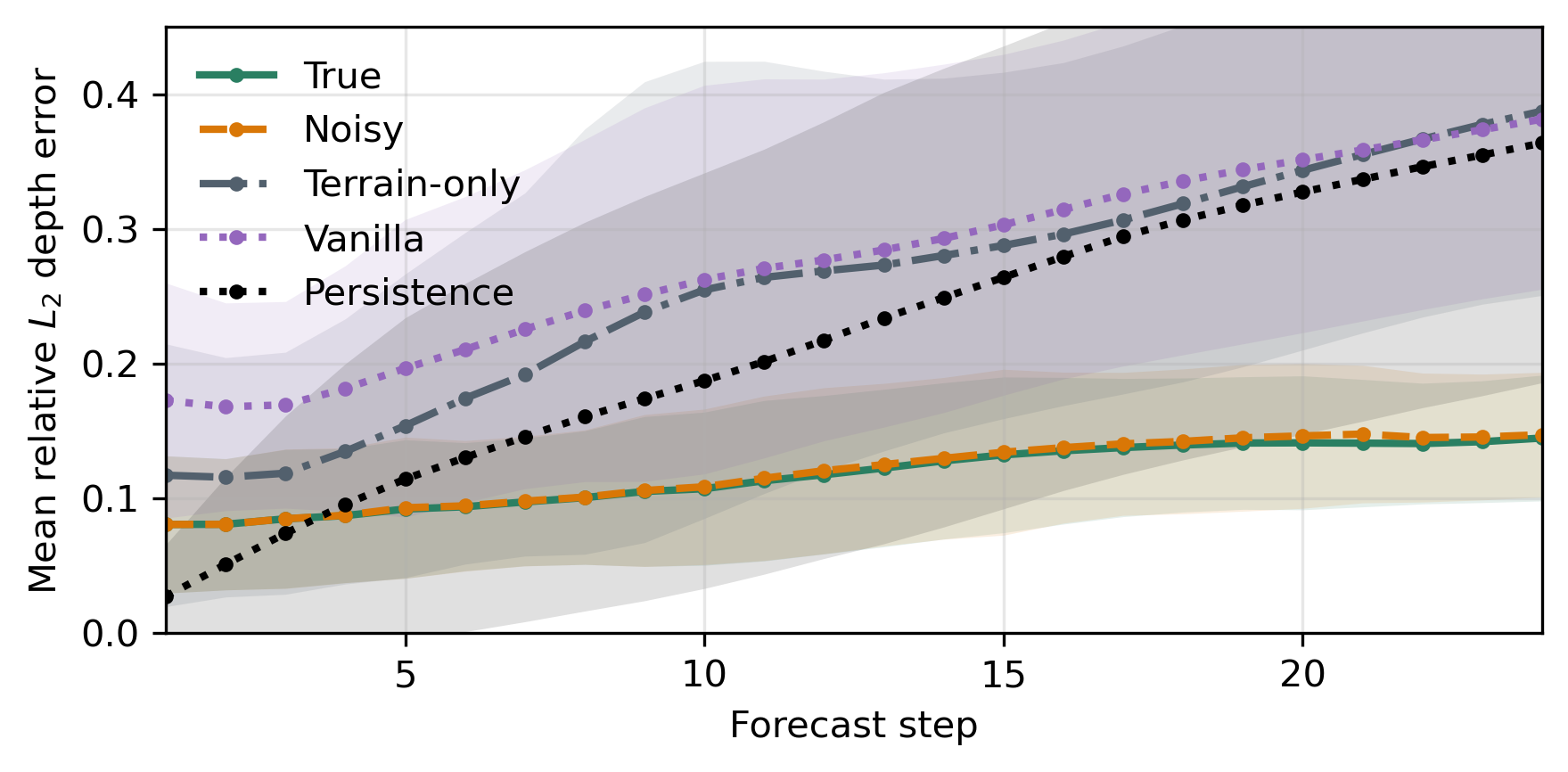}
\vspace{-0.3em}
\caption{Relative depth error over a 24-step forecast. ``True'' and ``Noisy'' denote full C-STRIDE with observed and perturbed future rainfall, respectively. Terrain-only and vanilla receive no rainfall. Persistence holds the complete preceding reference field fixed. Curves show means over the forecasts, with shading denoting one standard deviation. Forecasts can overlap in time, so this spread does not represent uncertainty across independent training runs.}

\label{fig:future_precip}
\vspace{-0.5em}
\end{figure}

Rainfall mismatch increases the full-model horizon-$24$ error from $14.48\%$ to $14.72\%$, approximately $0.24$ percentage points (Fig.~\ref{fig:future_precip}). Horizon one is unchanged because future rainfall enters after the first prediction. The model is therefore only weakly sensitive to this type of perturbation. The true- and noisy-rainfall curves stay close throughout, whereas the configurations without rainfall show much larger error growth; the tested perturbation matters far less than omitting rainfall altogether.

\subsection{Sensitivity to the observation interface}
\label{sec:ablation_results}

We test two properties of the observation interface: how much temporal history is needed to identify the flood state, and how performance changes when models are trained and evaluated with missing or noisy inputs.

\subsubsection{Window length}
\label{sec:ablation_window}

\begin{figure}[ht]
\centering
\includegraphics[width=0.65\columnwidth]{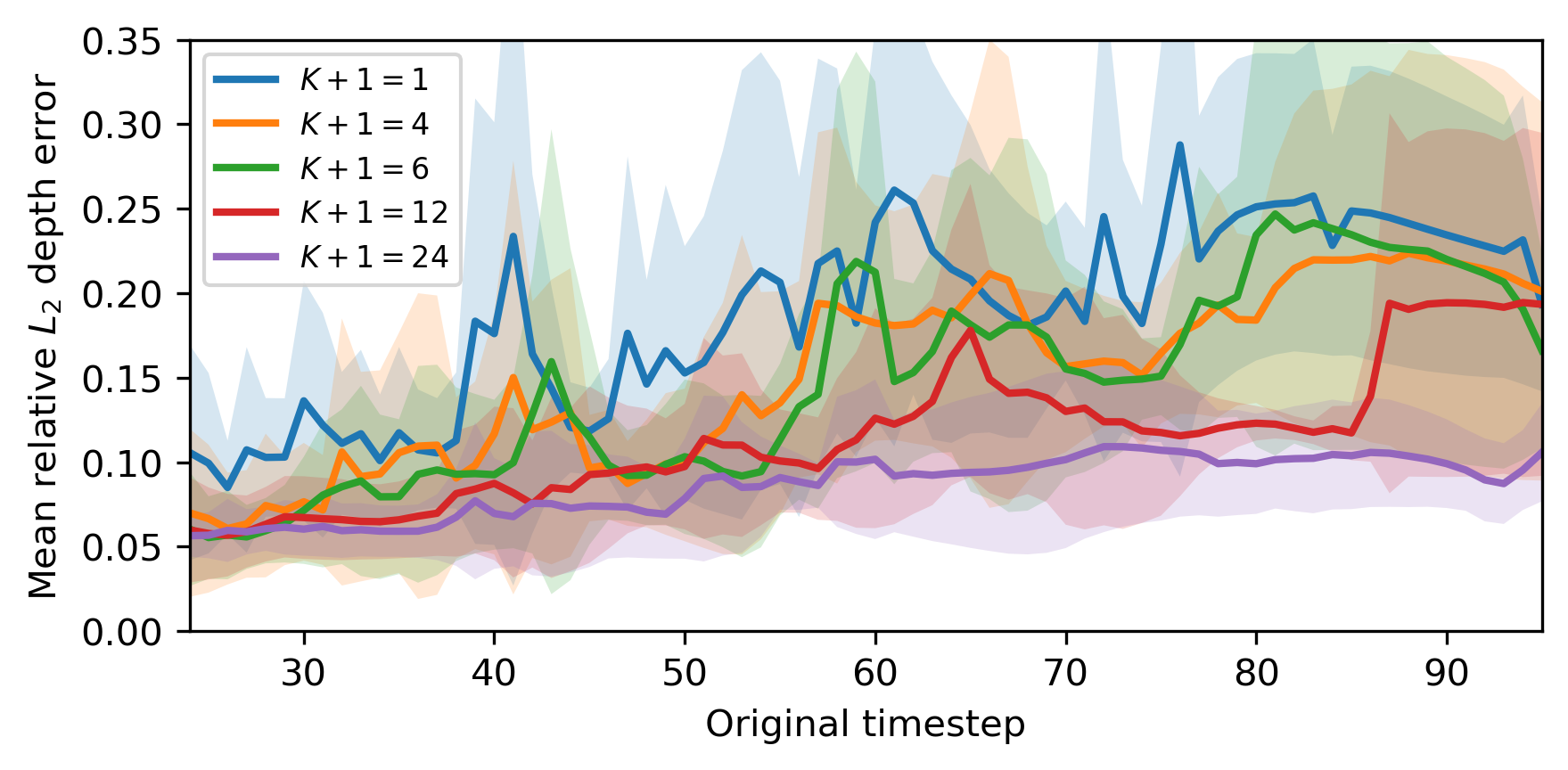}
\vspace{-0.3em}
\caption{Effect of observation-history length $K+1$ on full C-STRIDE prediction. The models are compared at common prediction times, so each curve covers the same part of the flood evolution. Curves show mean snapshot relative $L_2$ depth error, and shading denotes one standard deviation across events.}

\label{fig:ablation_lag}
\vspace{-0.5em}
\end{figure}

For window lengths $K+1=1,4,6,12,24$, mean relative errors on common prediction times are $18.72\%$, $15.15\%$, $14.69\%$, $11.44\%$, and $8.61\%$, respectively (Fig.~\ref{fig:ablation_lag}). Corresponding RMSEs are $0.1134$, $0.0929$, $0.0901$, $0.0701$, and $0.0508$\,m. Both metrics decrease steadily with history length, and the $24$-step history outperforms the $12$-step default. Comparing at common prediction times ensures that the differences reflect the available history rather than the part of the event evaluated. We retain the $12$-step window as the default for consistency with the main experiments; the $24$-step result indicates that longer histories could further improve accuracy, at the cost of a longer record before the first prediction.

\subsubsection{Sensor robustness}
\label{sec:robustness}

Table~\ref{tab:sensor_robust} compares sensor configurations, each with its own trained model. In the $50\%$ dropout configuration, the four mainstem gauges are always available, while the two tributary records are removed together at selected times and filled with the nearest available observation in time. The four-gauge configuration omits the tributary gauges entirely. The noise configurations add zero-mean Gaussian noise to the normalized inputs during training and evaluation. At each time step, its standard deviation is the noise level times the standard deviation across the six gauge values for gauge inputs, and across the $507$ rainfall values for rainfall inputs.

\begin{table}[!htb]
\centering
\small
\setlength{\tabcolsep}{5pt}
\renewcommand{\arraystretch}{1.15}
\begin{tabular}{lcc}
\toprule
\textbf{Configuration} & $\varepsilon_h$ (\%) & $\mathrm{RMSE}_h$ (m) \\
\midrule
Full, six gauges & $10.63$ & $0.0631$ \\
$50\%$ tributary dropout & $11.15$ & $0.0665$ \\
Four mainstem gauges & $10.67$ & $0.0636$ \\
Noise level $0.05$ & $11.77$ & $0.0700$ \\
Noise level $0.10$ & $12.17$ & $0.0730$ \\
Noise level $0.20$ & $12.43$ & $0.0741$ \\
\bottomrule
\end{tabular}
\caption{Mean relative depth error and RMSE for the sensor configurations.}
\label{tab:sensor_robust}
\end{table}

Using four mainstem gauges gives nearly the same error as the six-gauge model ($10.67\%$ versus $10.63\%$), while tributary dropout increases it to $11.15\%$. Error rises with noise level, reaching $12.43\%$ at level $0.20$, approximately $17\%$ above the default. The near-identical four- and six-gauge averages contrast with the mixed results at the held-out tributaries: similar field-level accuracy can hide local differences. Accuracy degrades gradually as the noise level increases.

\subsection{Computational cost}
\label{sec:efficiency}

Training the full encoder--decoder took approximately $26.2$\,h, a one-time cost. On one NVIDIA L40S GPU, next-step prediction of all fields of the April 2013 event takes $21.99$\,s, or approximately $0.262$\,s per field. Restricting the queries to the six gauge locations reduces this to $3.55$\,ms per snapshot. The $40$ forecasts of $24$ steps take $247.37$\,s in total, including forecast preparation and scoring.

For reference, the reported CLDNet surrogate evaluation time is $28.8$\,s per event~\citep{si_toward_2026}, while SynxFlow requires approximately $3{,}300$\,s to simulate a $96$-h event. These measurements have different scopes: SynxFlow advances depth and momentum over the full computational grid, whereas the learned models predict depth at selected locations, and the C-STRIDE forecast timing covers multiple forecast origins with overlapping target times. Because SynxFlow was also run on an NVIDIA L40S GPU, next-step prediction of the April 2013 event ($21.99$\,s) is approximately $150$ times faster than the simulation, subject to these differences in scope.

%% file: sec_discussion.tex
\section{Discussion}
\label{sec:discussion}

\subsection{What gauges, terrain, and precipitation each contribute}
\label{sec:disc_roles}

The model variants separate three information sources with distinct roles. Terrain conditioning gives the largest single gain, about half of the total reduction in four-event next-step error achieved by the full model ($18.14\%\rightarrow14.23\%$, compared with $18.14\%\rightarrow10.63\%$), and most of the improvement in inundation extent for the 2013 event. Two explanations, not mutually exclusive, are plausible. Elevation, slope, and roughness carry physical information about where water collects and how fast it moves. They also give the decoder high-resolution positional information: with only two coordinates, a pointwise decoder must represent detail on a $5{,}075\times1{,}661$ raster, whereas terrain features vary on the $30$\,m grid itself, and the binary Manning feature marks channels and open water directly. Removing terrain features one at a time, and giving vanilla STRIDE a richer coordinate encoding, would distinguish these explanations. This distinction matters when assessing whether the learned terrain relationships transfer to other basins.

Precipitation alone gives a smaller next-step improvement than terrain, but its contribution becomes more pronounced at longer horizons. Added to terrain, it still reduces the four-event next-step error from $14.23\%$ to $10.63\%$. At the next-step horizon, the gauges already record the hydraulic response to past rainfall, so rainfall adds information mainly in parts of the basin that the gauges do not see. For forecasting, the auxiliary model cannot anticipate new rainfall from gauge histories alone, and the configurations without precipitation accumulate substantially larger errors over $24$\,h. The cost of precipitation input appears at the gauges, where full C-STRIDE has larger peak errors than the terrain-only model, particularly at the tributaries. Because the $507$ precipitation features outnumber the $6$ gauge features, compressing rainfall before it enters the encoder, for example into sub-catchment averages, is a natural next step.

Gauge histories provide the state information. A single snapshot is substantially less informative than a sequence. The larger late-event errors with shorter windows suggest that longer histories help distinguish evolving flood states, consistent with the delay-observability motivation in Section~\ref{sec:stride_recap}. Removing the tributary gauges changes basin-wide error only slightly, which shows that aggregate metrics are not sufficient for judging a sensor network; local objectives, such as hydrograph fidelity on the tributaries, should be evaluated separately.

\subsection{From simulated to observed gauge inputs}
\label{sec:disc_sim2real}

C-STRIDE is trained only on simulated gauge histories, so USGS records test how it behaves when the inputs differ systematically from those seen in training. These differences arise from simulator errors in timing and recession, visible where SynxFlow departs from the USGS hydrographs; from the conversion of gauge stage to model depth, which depends on the station datum and on how the DEM represents the channel bed; from the initial state of each simulation; and from measurement noise. The USGS-input experiment shows that the learned state responds to observed timing and magnitude at several gauges without retraining. It also shows that such inputs can produce artifacts, such as the spurious early rise at Russell and the crest overshoot at DuPage. At the two tributary gauges withheld from the inputs (Fig.~\ref{fig:usgs_heldout}), the response is mixed: observed mainstem histories raise NSE at Salt Creek from $0.004$ to $0.346$, although its peak remains $7$\,h late, but leave DuPage essentially unchanged. Fine-tuning on observed histories, training with perturbed or bias-shifted inputs, and estimating datum offsets from pre-event baseflow are direct ways to narrow this gap.

\subsection{Toward an operational flood digital twin}
\label{sec:disc_operational}

In the pillar framework of Section~\ref{sec:motivation}, C-STRIDE supplies fast spatial prediction (P2) and observation-driven state estimation (P3) and inherits physical anchoring (P1) from SynxFlow. Several elements of an operational digital twin remain to be added.

\paragraph{Operational forcing.}
Long-horizon skill depends on future rainfall. The perturbation experiment tests one specific error model. Operational precipitation forecasts also have storm-displacement, timing, and systematic intensity errors, as well as missed and false storms. Prospective evaluation with real-time gauge streams and archived precipitation forecasts or ensembles is the most important next test. Independent inundation observations, including remotely sensed flood extents, would complement the gauge comparisons and assess spatial accuracy beyond agreement with SynxFlow.

\paragraph{Uncertainty and data assimilation.}
C-STRIDE produces deterministic estimates. The spread across evaluation samples in the figures describes variability in performance, not predictive uncertainty for an individual forecast. The compact latent state could initialize a latent ensemble filter such as LD-EnSF~\citep{xiao2026ldensf} or a latent ensemble-variational smoother such as LEVDA~\citep{levda2026}, with the coordinate decoder acting as the observation operator at gauge locations, subject to a suitable measurement-error model. Combined with precipitation ensembles and estimates of surrogate error, this would support probabilistic forecasts of depth and extent. A complete operational twin would additionally require a decision interface and feedback to the physical system~\citep{nasem2024digitaltwins}.

\paragraph{Changing sensor networks.}
The current encoder assumes a fixed set of gauges. An encoder that represents sensor locations and availability explicitly, as in attention-based sparse-sensing models~\citep{santos2023senseiver}, could accommodate outages and new sensors without retraining. Future tests should distinguish isolated missing values, contiguous outages, and permanent sensor loss, using only information available at each forecast origin. Sensor-placement studies could then examine where additional measurements most improve local hydrograph fidelity or inundation extent.

\paragraph{Computational role.}
Because SynxFlow already runs faster than real time for single events, the surrogate's advantage lies in workloads with many evaluations: ensembles, assimilation cycles, and scenario analysis. The measured inference costs support these repeated-use settings, but a break-even estimate would require matched computational scopes and an accounting of both training-data generation and model training (Section~\ref{sec:efficiency}).

\paragraph{Transfer and broader outputs.}
Multi-basin training and basin-conditioned encoders could test whether terrain-aware decoding reduces the data needed to adapt to a new watershed. Such studies should distinguish transfer across terrain, storm regimes, and sensor layouts. Extending the predicted state to velocity and discharge, together with conservation-aware training, would support hazard measures beyond depth and inundation extent. Coupling the surrogate with upstream hydrologic predictions would allow evaluation under time-varying inflow conditions beyond the precipitation-driven experiments presented here.

\subsection{Limitations}
\label{sec:limitations}

The evaluation covers one basin, four held-out test events, and a six-gauge network, together with a retrained four-gauge configuration. Each configuration was trained once, so small differences between configurations may fall within run-to-run variation. The missing-observation experiment uses a separately trained model with $50\%$ dropout at two tributary gauges and nearest-neighbor temporal filling. Such filling may use later observations, so it does not establish robustness to causal, real-time outages. The reduced-network experiment requires retraining and does not demonstrate adaptation to arbitrary sensor layouts or permanent gauge loss during deployment. Field metrics are evaluated on event-specific masks derived from reference depths, not on the full computational domain. These masks select cells that reach $0.1$\,m in each event and would not be known in advance during deployment; accuracy outside them is not established.

The surrogate learns water depth from simulated targets and inherits the simulator's limitations: the representation of terrain and channel geometry at $30$\,m, two-class Manning roughness, and the initial and boundary conditions of each event. It can also inherit errors in gauge-to-grid alignment. The USGS comparisons provide a complementary observational assessment, but the WSE comparisons aligned using event-wide mean offsets do not establish absolute elevation accuracy or prospective deployment performance. Evaluation at gauges that also supply the input histories is an observation-consistency test; the held-out-gauge experiment covers only two tributary gauges in one event. Training on a physical simulator does not explicitly enforce mass conservation or hydraulic consistency in the learned predictions, and velocity and discharge are outside the present scope.

Forecasts remain conditional on precipitation information. The mismatch experiment tests a specific perturbation of the reference rainfall rather than the full range of errors in operational precipitation forecasts, such as storm displacement, timing errors, and persistent bias. The model also produces deterministic estimates without calibrated predictive uncertainty.

%% file: sec_conclusion.tex
\section{Conclusions}
\label{sec:conclusion}

We presented Conditional STRIDE (C-STRIDE), an observation-driven AI digital twin for basin-wide flood prediction from sparse stream gauges. A recurrent encoder summarizes gauge and, optionally, precipitation histories into a latent basin state, and a terrain-conditioned coordinate decoder predicts water depth one step beyond the observation window. A separately trained gauge forecaster supports multi-step prediction when new observations are unavailable. Trained on SynxFlow simulations of the Des Plaines River basin, the model combines rapid spatial prediction with observation-driven state estimation, while inheriting its physical reference from the simulator.

The experiments support four conclusions.
\begin{enumerate}
    \item Terrain and precipitation provide complementary information for next-step prediction. Terrain gives the larger individual gain, while full conditioning reduces four-event mean relative depth error from $18.14\%$ for vanilla STRIDE to $10.63\%$. For the April 2013 flood, full C-STRIDE achieves CSI $0.933$ at the $0.5$\,m depth threshold on the assessed domain.
    \item Precipitation limits error growth during multi-step prediction. With reference future rainfall, full C-STRIDE reaches $14.48\%$ mean relative error at $24$\,h, compared with $38.14\%$ for vanilla STRIDE. The tested rainfall mismatch raises the full-model error to $14.72\%$. Full-field persistence is more accurate at the first step, whereas C-STRIDE is more accurate at the reported horizons from four steps onward.
    \item Field accuracy and gauge fidelity are distinct objectives. Full conditioning gives the lowest aggregate field error, whereas terrain-only conditioning gives the best average hydrograph agreement with SynxFlow at the six gauges. Similar field errors for four- and six-gauge configurations can therefore conceal local differences.
    \item Predictions respond to observed histories without retraining, but the gains are site dependent. Replacing simulated histories with USGS observations improves NSE at three of six gauges, including an increase from $0.118$ to $0.840$ at Riverside. The mixed response at the two tributary gauges withheld from encoder inputs further limits claims of general observational correction. These comparisons use gauge-specific, event-wide mean offsets and assess the retrospective response to observed inputs.
\end{enumerate}

Full C-STRIDE predicts the April 2013 depth fields in $21.99$\,s on one NVIDIA L40S GPU, about $150$ times faster than SynxFlow on the same hardware. The results are limited to one basin, event-specific reference-derived evaluation grids, and deterministic predictions. Priorities for further development are prospective tests with operational precipitation forecasts and causal gauge preprocessing, independent inundation observations, calibrated uncertainty, and evaluation across basins and sensor networks.

\subsection*{Data and code availability}

{\sloppy Code, input data, and trained models for C-STRIDE will be available at \url{https://github.com/Flood-Digital-Twins/CSTRIDE}, organized as in the CLDNet repository for the same Des Plaines River basin dataset (\url{https://github.com/Flood-Digital-Twins/CLDNet}; \citealp{si_toward_2026}). The simulated flow fields will not be hosted but can be regenerated from the released inputs with the open-source SynxFlow solver~\citep{synxflow_software}. USGS gauge records were obtained from the National Water Information System~\citep{usgs_nwis}.\par}

\subsection*{Declaration of competing interest}

The authors declare that they have no known competing financial interests or personal relationships that could have appeared to influence the work reported in this paper.

\subsection*{Acknowledgments}

We thank Dr.~Omar Sallam (Argonne National Laboratory) for calibrating the SynxFlow model of the Des Plaines River basin, and Dr.~Eugene Yan (Argonne National Laboratory), Prof.~Barnali Dixon (University of South Florida), and Prof.~Subhrajit Guhathakurta (Georgia Institute of Technology) for helpful discussions. This work was supported by the National Science Foundation under Awards CNS-2325631 and DMS-2245111, by the U.S. Department of Energy under Contract Nos. DE-AC05-00OR22725 and DE-AC02-06CH11357, and by the 2025 IDEaS + Cloud Hub with support from Microsoft at the Georgia Institute of Technology.

\subsection*{Declaration of generative AI and AI-assisted technologies in the writing process}

During the preparation of this work, the authors used Claude and ChatGPT to assist with manuscript editing, table revision, and language polishing. After using these tools, the authors reviewed and edited the content as needed and take full responsibility for the content of the publication.